\documentclass[sigconf]{acmart}

\AtBeginDocument{%
  \providecommand\BibTeX{{%
    \normalfont B\kern-0.5em{\scshape i\kern-0.25em b}\kern-0.8em\TeX}}}

\setcopyright{acmcopyright}

\copyrightyear{2022}
\acmYear{2022}
\setcopyright{rightsretained}
\acmConference[MM '22]{Proceedings of the 30th ACM International Conference on Multimedia}{October 10--14, 2022}{Lisboa, Portugal}
\acmBooktitle{Proceedings of the 30th ACM International Conference on Multimedia (MM '22), Oct. 10--14, 2022, Lisboa, Portugal}
\acmISBN{978-1-4503-9203-7/22/10}
\acmDOI{10.1145/3503161.3548282}

\usepackage{etoolbox}
\makeatletter
\patchcmd{\maketitle}{\@copyrightpermission}{
   \begin{minipage}{0.3\columnwidth}
     \href{https://creativecommons.org/licenses/by/4.0/}{\includegraphics[width=0.90\textwidth]{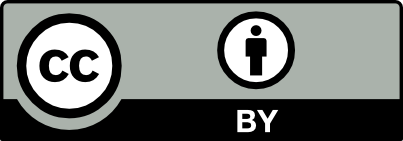}}
   \end{minipage}\hfill
   \begin{minipage}{0.7\columnwidth}
     \href{https://creativecommons.org/licenses/by/4.0/}{This work is licensed under a Creative Commons Attribution International 4.0 License.}
   \end{minipage}
   \vspace{5pt}
}{}{}

\makeatother

\acmPrice{15.00}

\newcommand{\final}{1}
\usepackage{ifthen}
\usepackage{xcolor}

\definecolor{WeimingColor}{rgb}{0,0,0.8}
 
\definecolor{FanColor}{rgb}{0.8,0,0.8}

\newcommand{\weiming}[1]{{\color{WeimingColor} [Weiming: #1]}}

\newcommand{\fan}[1]{{\color{FanColor}[Fan: #1]}}

\newcommand{\warning}[1]{{\it\color{red} #1}}
\newcommand{\toremove}[1]{{\it\color{red} (To remove) #1}}
\newcommand{\note}[1]{{\it\color{blue} #1}}
\newcommand{\nothing}[1]{}

\ifthenelse{\equal{\final}{1}}
{
\renewcommand{\weiming}[1]{}
\renewcommand{\fan}[1]{}

\renewcommand{\warning}[1]{}
\renewcommand{\toremove}[1]{}
\renewcommand{\note}[1]{}
\renewcommand{\nothing}[1]{}
}{}

\usepackage{algorithm}
\usepackage{balance}
\usepackage{algorithmic}

\begin{document}
\title{Draw Your Art Dream: Diverse Digital Art Synthesis with Multimodal Guided Diffusion}

\author{Nisha Huang}
\email{huangnisha2021@ia.ac.cn}
\affiliation{%
  \institution{School of Artificial Intelligence, UCAS}
  \institution{NLPR, Institute of Automation, CAS}
  \country{China}
}

\author{Fan Tang}
\email{tfan.108@gmail.com}
\orcid{0000-0002-3975-2483}
\affiliation{%
  \institution{School of Artificial Intelligence, Jilin University}
  \country{China}
}

\author{Weiming Dong$^*$}
\email{weiming.dong@ia.ac.cn}
\affiliation{
\institution{NLPR, Institute of Automation, CAS}
\institution{School of Artificial Intelligence, UCAS}
\thanks{Corresponding author}
\country{China}
}

\author{Changsheng Xu}
\email{csxu@nlpr.ia.ac.cn}
\affiliation{
\institution{NLPR, Institute of Automation, CAS}
\institution{School of Artificial Intelligence, UCAS}
\country{China}
}

\renewcommand{\shortauthors}{Nisha Huang, Fan Tang, Weiming Dong, \&amp; Changsheng Xu}
\renewcommand{\authors}{Nisha Huang, Fan Tang, Weiming Dong, and Changsheng Xu}
\begin{teaserfigure}
  \includegraphics[width=\textwidth]{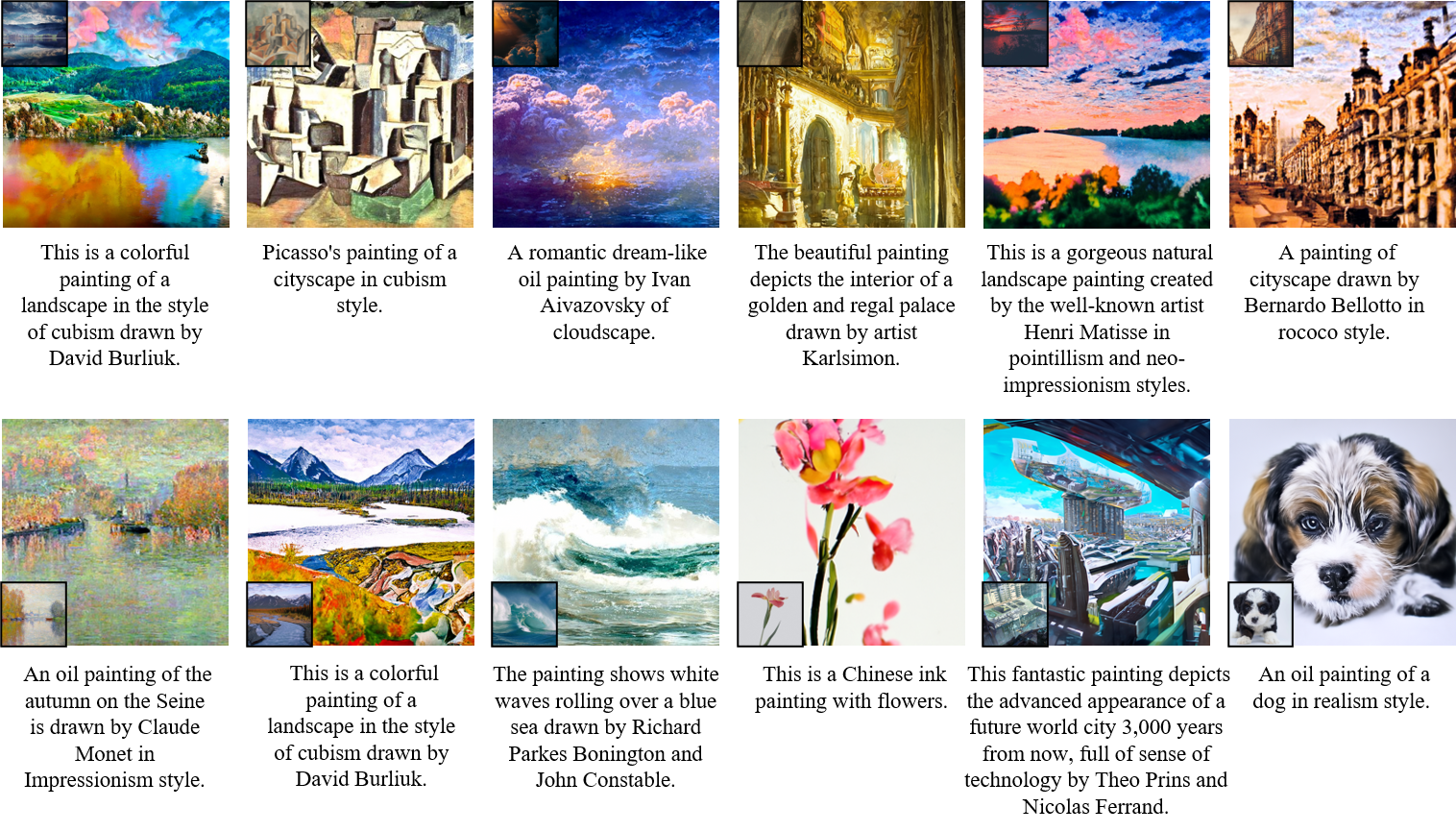}
  \vspace{-7mm}
  \caption{Digital art paintings generated by the proposed multimodal guided artwork diffusion (MGAD) model.}
  \label{fig:teaser}
\end{teaserfigure}
\begin{abstract}
Digital art synthesis is receiving increasing attention in the multimedia community because of engaging the public with art effectively. 
Current digital art synthesis methods usually use single-modality inputs as guidance, thereby limiting the expressiveness of the model and the diversity of generated results. 
To solve this problem, we propose the multimodal guided artwork diffusion (MGAD) model, which is a diffusion-based digital artwork generation approach that utilizes multimodal prompts as guidance to control the classifier-free diffusion model.
Additionally, the contrastive language-image pretraining (CLIP) model is used to unify text and image modalities. 
Extensive experimental results on the quality and quantity of the generated digital art paintings confirm the effectiveness of the combination of the diffusion model and multimodal guidance. 
Code is available at \url{https://github.com/haha-lisa/MGAD-multimodal-guided-artwork-diffusion.}
\end{abstract}



\maketitle

\section{Introduction}
\label{sec:intro}

Many people like to appreciate paintings, but not everyone has the expertise and labor time to create an ideal artwork.
Therefore, a tool that can create paintings with high quality and a wide variety from simple inputs would be helpful for novice people to experience art creation easily.
In recent years, research on this topic has drawn extensive attention because of its scientific and artistic values.

Existing works use computational algorithms~\cite{Wang:2014:TPW,Zhang:2016:GKA,Alvarez:2021:IDR} or style transfer approaches~\cite{johnson2016perceptual,Huang:2017:AdaIn,liu2021adaattn,deng2021stytr2} for art creation.
The task of style transfer involves transferring the style of an image or a collection of images (a pictorial artwork or the artworks of a painter) to another image (usually a photograph) or video clip.
Style transfer~\cite{deng2020arbitrary,deng2021arbitrary,Wei:2022:ACS,Zhang:2022:CAST} can generate high-quality art images/videos but does not closely resemble the actual paintings. The resulting content depends on the content of the input photograph, hence limiting the controllability of the creation process and the diversity of the results.
Image-to-image translation scheme was also used for artistic image generation~\cite{Zhu:2017:CycleGAN,kotovenko2019content,Lin:2021:DAM}. 
Nevertheless, these methods can only generate results with the style of one or a limited number of artists.
Diversified style transfer methods~\cite{Wang:2020:DAS,Chen:2021:DIS,chen2021dualast} were proposed to generate multiple results from the same input, but all the contents of the results are still the same as the input content image.
Recently, some new works used text to control the generation of paintings~\cite{clipdraw, styleclipdraw, vector}, based on dual language-image encoders, such as CLIP~\cite{openai}.
However, the quality and diversity of the paintings produced by the text guidance still need to be improved urgently. Therefore, we strive to create user-controlled, realistic, high-quality, and diverse artworks.

Meanwhile, outstanding works in image generation have been produced~\cite{AdityaRamesh2021ZeroShotTG,BoZhao2018MultiViewIG,TaoXu2018AttnGANFT,ruan2021dae,huo2021survey,xue2022deep}. However, few of them were employed to generate digital paintings. Diffusion models that emerged lately~\cite{VQ-Diffusion,latentDiffusion,Guided-Diffusion} achieve state-of-the-art quality and outstanding diversity, and have the potential to create excellent digital artworks. Restricted by the guidance conditions of existing diffusion models~\cite{DDPM,DDIM}, generating digital artworks freely is challenging. For example, ADM-G~\cite{Guided-Diffusion} can only generate natural images of a certain category based on the category label. Therefore, CLIP~\cite{openai}, which enables multimodal prompts as guidance conditions to broaden the application of diffusion models to generate digital artworks, was used. A freshly proposed form of guidance known as classifier-free guidance~\cite{ho2021classifier} could produce similar results without the use of a separate classifier. Therefore, we combine the classifier-free diffusion model~\cite{ho2021classifier} and CLIP~\cite{openai} for digital art generation, which has the advantage of high quality and superb diversity.

\begin{figure}[!t]
    \centering
    \includegraphics[width=8.5cm]{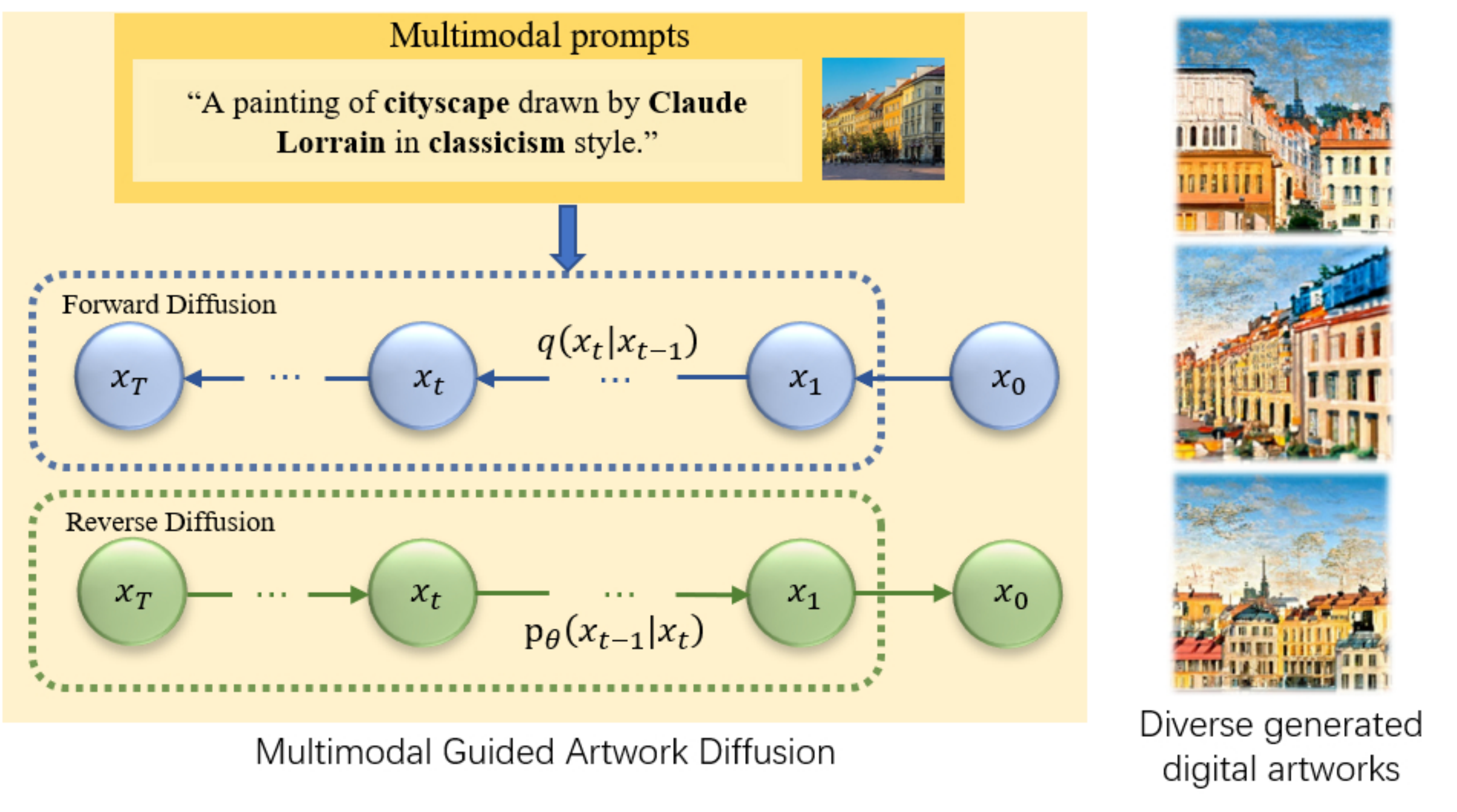}
    \caption{Insight into our work. The diverse generated digital artworks are guided by CLIP loss via the classifier-free diffusion model.}
    \label{fig:insight}
\end{figure}

Taking advantage of the capabilities of guided diffusion models to generate images and the abilities of text-to-image or image-to-image models to handle prompts, guided diffusion is applied to the issue of multimodal-conditional digital artwork synthesis. 
In this paper, we propose a multimodal guided artwork diffusion (\textbf{MGAD}) model, which adopts \textbf{CLIP} to assist multimodal prompts for generating digital artworks guided by classifier-free diffusion models (Figure~\ref{fig:insight}).
We begin by utilizing a fine-tuned unconditional model with $512 \times 512$ resolution for the diffusion model from OpenAI's class-conditional ImageNet diffusion model~\cite{Guided-Diffusion}. Then, we employed a secondary diffusion model~\cite{model} that was previously trained on the Yahoo Flickr Creative Commons 100 million~\cite{yfcc100} dataset to produce higher-quality outputs. Findings show that the samples (Figure~\ref{fig:teaser}) from our model generated pleasing and artistic results.
The main contributions of this work are summarized as follows:
\begin{itemize}
\item We propose MGAD, a method used to refine each transition in the generative process by matching each latent variable with the given multimodal guidance.
\item We enable the user to control the semantic content of the prompts with respect to the similarity of the results by using CLIP and the classifier-free diffusion model.
\item The aggregate experimental results show that MGAD outperforms the baseline approaches and achieves excellent results in terms of diversity and quality of the generated digital artworks.
\end{itemize}

\section{RELATED WORK}
\paragraph{Guided Image Synthesis}
Previous research has focused on the combination of natural language and image, including tasks, such as text-guided~\cite{YahuiLiu2020DescribeWT,DiffusionCLIP} or image-guided synthesis~\cite{HaoWang2021CycleConsistentIG,ILVR}. 
Consequently, some prominent works for vision and language–\ representations~\cite{desai2021virtex,chen2020uniter,li2020oscar} have been studied in depth to integrate image–text embedding.
The pre-trained text–image embedding model CLIP~\cite{openai} can be efficiently transferred non–trivially to most tasks, generally without any specific dataset training compared with fully supervised baselines. 
Its representations have been proven robust and comprehensive enough to perform zero-shot classification and various vision-language tasks on different datasets. The combination of CLIP and GANs, which utilize CLIP to guide the optimization of a latent code for a required image manipulated or generated, has emerged in subsequent studies. For instance, StyleCLIP~\cite{StyleCLIP} used CLIP embedding vectors to tune the latent codes. StyleGAN-NADA~\cite{StyleGANNADA} utilized CLIP frame to adjust the zero-shot domain.
Our goal is to recognize these descriptions automatically.
CLIPstyler~\cite{CLIPstyler} proposed patchCLIP for transferring semantic texture information on text conditions. 

GLIDE~\cite{glide} and DALL-E 2~\cite{dalle2} focus on open domain image synthesis. 
Both of them are implemented with the idea of integrating image generators and joint text-image encoders into their architectures. 
They all contain pre-trained models with large-scale datasets of numerous text-image pairs. 
In contrast, we do not intend to spend such expensive training resources and time. 
We aim to only use CLIP to measure the similarity between the prompts and the generated results to guide the reverse direction.
In addition, the above CLIP-based image editing methods only allowed the users to supply a textual description as the style condition.
To improve controllability, we attempt to input multiple modality information as control conditions for art image synthesis.

\paragraph{Diffusion Models}
Diffusion models~\cite{2015Diffusion}, which consist of one forward process (signal to noise) and one reverse process (noise to signal), have been newly demonstrated to produce high-quality images~\cite{JaschaSohlDickstein2015DeepUL,song2019generative,DDPM,Guided-Diffusion}. 
Denoising diffusion probabilistic models (DDPM)~\cite{DDPM} and score-based generative models~\cite{song2019generative,YangSong2021ScoreBasedGM}, have recently gained remarkable success in the field of image generation~\cite{DDPM,DDIM,YangSong2021ScoreBasedGM,AlexiaJolicoeurMartineau2021AdversarialSM}. 
Ablated diffusion model (ADM)~\cite{Guided-Diffusion} has demonstrated higher image synthesis quality than variational autoencoders (VAEs)~\cite{razavi2019generating}, flow-based models~\cite{DiederikPKingma2018GlowGF}, auto-regressive models~\cite{JacobMenick2018GeneratingHF}, and GANs~\cite{GAN,StyleGAN,StyleGAN2}. 
The generative power of these models~\cite{Guided-Diffusion,DDPM,YangSong2021ScoreBasedGM} stems from a natural adaptation to the inductive biases of image-like data when their underlying neural skeleton is implemented as a U-Net~\cite{UNet}.
Recently, diffusion models have been explored for conditional generation, such as class-conditional generation~\cite{ILVR}, image-guided synthesis~\cite{Guided-Diffusion}, text-guided synthesis~\cite{VQ-Diffusion,DiffusionCLIP}, semantics-guided synthesis~\cite{MoreCF}, and super-resolution~\cite{latentDiffusion}.

\paragraph{Art Painting Synthesis}
Artwork analysis~\cite{deng2020exploring,hall2015cross,deng2019selective} and synthesis~\cite{tang2017animated,huang2022dualface} are the most challenging tasks that enable effective engagement of the public with art, balancing between sophisticated computational/engineering techniques and artistic purposes. 
Tan et al.~\cite{ArtGAN, ImprovedArtGAN} proposed ArtGAN, where the label information was propagated back to the generator for more efficient learning. 
More recently, CLIPDraw~\cite{clipdraw} and StyleCLIPDraw~\cite{styleclipdraw} began generating paintings from randomized B\'ezier curves that fit a given text and style. 
In contrast, these two models mainly capture larger features, such as shapes or outlines, rather than fine-grained textures. 
Yi et al.~\cite{PaintingDiffusion} explored the generation of fine art painting that used diffusion models without any guidance.

Inspired by the aforementioned works, we propose a novel MGAD model, which concentrates on synthesizing drawings rather than realistic pictures.
We explore whether ADM~\cite{Guided-Diffusion} can be guided by text and image prompts to synthesize high-quality and fantasy art paintings rather than only using text prompts, which makes generating results less controllable.

\section{METHODS}
\subsection{Overview}
\begin{figure}[tbp]
    \centering
    \includegraphics[width=8.5cm]{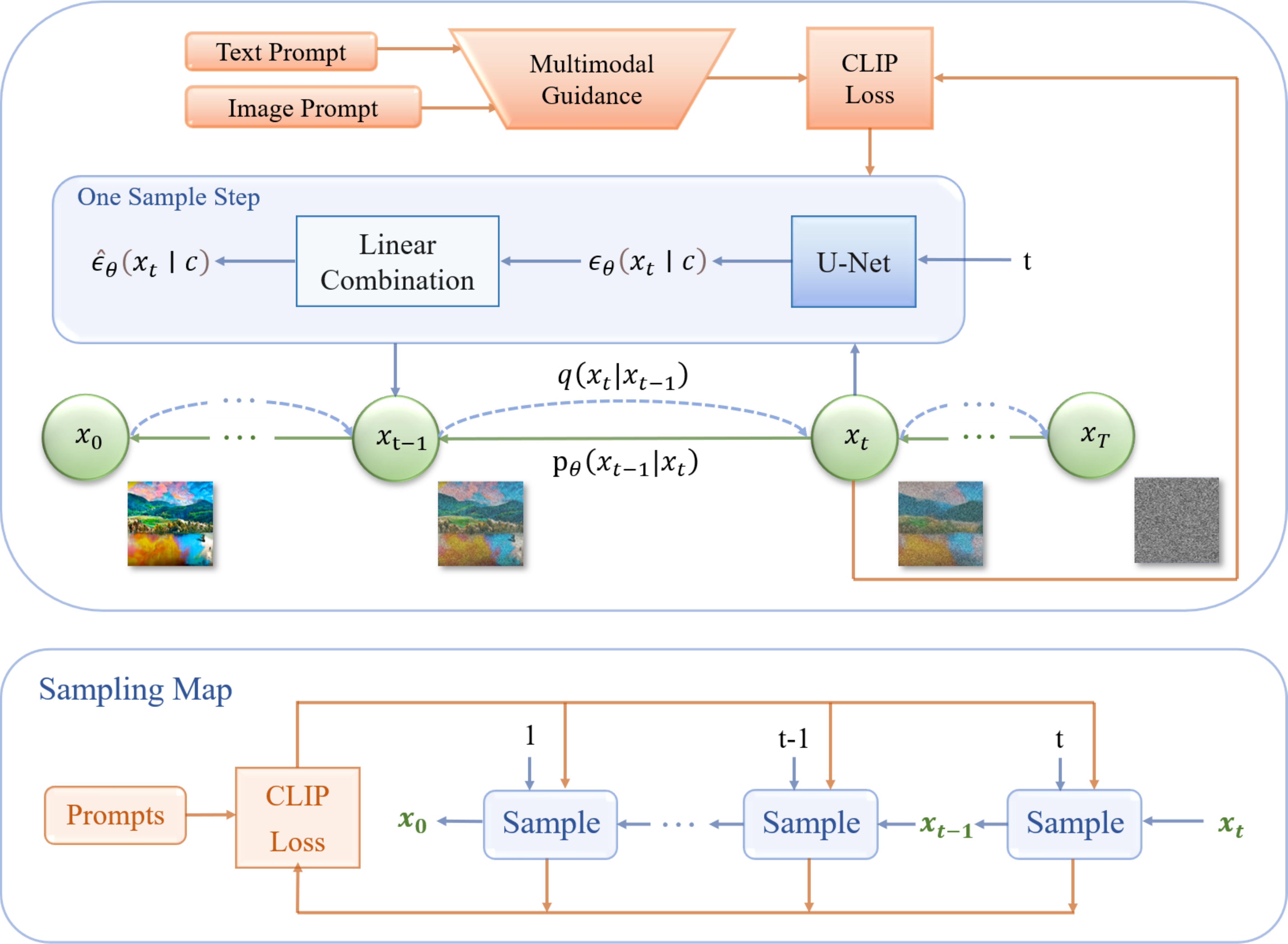}
    \caption{Overall framework of our method. One sample step consists of a U-Net and linear combination. Every sample step is guided by multimodal guidance. }
    \label{fig:pipeline}
\end{figure}
Figure~\ref{fig:pipeline} shows the overall framework of the proposed MGAD for art painting synthesis. MGAD is a new unified framework that incorporates different modality guidance into a pre-trained classifier-free diffusion model. 
Our goal is to generate the required art paintings according to one or both images and text prompts to take advantage of multimodal guidance. 
The multimodal guidance enables the controllable art painting synthesis and realizes complementarity between modes.
First, we utilize a pre-trained diffusion model $\epsilon_{\theta}$ to transform the noise image $x_{0}$ into latent $x_{t_{0}}\left(\theta\right)$. 
Second, the target $y_{tar}$ leads the diffusion model to generate samples guided by the CLIP loss at the reverse process. 
Third, the deterministic forward processes are based on DDPM~\cite{DDPM}. 
For translation among unseen domains, the image generation is also performed by combining multimodal guidance and diffusion models~\cite{Guided-Diffusion,ho2021classifier}.

\subsection{Multimodal Guided Diffusion}
Diffusion models~\cite{2015Diffusion} are inspired by non-equilibrium thermodynamics. 
They define a Markov chain of diffusion steps to add random noise to data slowly and then learn to reverse the diffusion process to construct the desired data samples from noise. 
Unlike VAEs~\cite{razavi2019generating} or flow-based~\cite{DiederikPKingma2018GlowGF} models, diffusion models are learned using a fixed procedure, and the latent variable has high dimensionality (same as the original data). 

A diffusion model includes one forward and one reverse diffusion process. 
The forward process is fixed to a Markov chain trained using variational inference, which gradually adds noise to the data. 
The data distribution is defined as $x_{0}\sim q\left(x_{0}\right)$ and a Markov chain forward process named as $q$. 
In addition, the data generates noised samples from $x_{1}$ to $x_{T}$. 
At each step of the forward process, Gaussian noise is added to the data accordingly by a variance schedule $\beta_{1},...,\beta_{T}$: 
\begin{equation}
q\left(\mathbf{x}_{1: T} \mid \mathbf{x}_{0}\right):=\prod_{t=1}^{T} q\left(\mathbf{x}_{t} \mid \mathbf{x}_{t-1}\right),
\end{equation}
\begin{equation}
\quad q\left(\mathbf{x}_{t} \mid \mathbf{x}_{t-1}\right):=\mathcal{N}\left(\mathbf{x}_{t} ; \sqrt{1-\beta_{t}} \mathbf{x}_{t-1}, \beta_{t} \mathbf{I}\right).
\end{equation}
The training goal is to optimize the negative log likelihood of the usual variational bound:
\begin{equation}
\begin{gathered}
L:=\mathbb{E}_{q}\left[-\log p\left(\mathbf{x}_{T}\right)-\sum_{t \geq 1} \log 
\frac{p_{\theta}\left(\mathbf{x}_{t-1} \mid \mathbf{x}_{t}\right)}{q\left(\mathbf{x}_{t} \mid \mathbf{x}_{t-1}\right)}\right], \\
=\mathbb{E}_{q}\left[-\log \frac{p_{\theta}\left(\mathbf{x}_{0: T}\right)}{q\left(\mathbf{x}_{1: T} \mid \mathbf{x}_{0}\right)}\right],\\
\leq \mathbb{E}\left[-\log p_{\theta}\left(\mathbf{x}_{0}\right)\right].
\end{gathered}
\label{equ:1}
\end{equation}
Forward process samples $\mathbf{x}_{t}$ at time step $t$:
\begin{equation}
q\left(x_{t} \mid x_{t-1}\right):=\mathcal{N}\left(x_{t} ; \sqrt{\alpha_{t}} x_{t-1},\left(1-\alpha_{t}\right), \mathcal{I}\right),
\end{equation}
where $\alpha_{t}:=1-\beta_{t}$ and $\bar{\alpha}_{t}:=\prod_{s=0}^{t} \alpha_{s}$. 
The variance of the noise for an arbitrary time step is defined as $1-\bar{\alpha}_{t}$ to determine the noise schedule. 
The form of the diffusion models~\cite{2015Diffusion} can be represented by $p_{\theta}\left(\mathbf{x}_{0}\right):=\int p_{\theta}\left(\mathbf{x}_{0: T}\right) d \mathbf{x}_{1: T}$. $p_{\theta}\left(\mathbf{x}_{0: T}\right)$ is the expression of the reverse process that is defined as a learning Gaussian distribution Markov chain that initiates at $p\left(\mathbf{x}_{T}\right)=\mathcal{N}\left(\mathbf{x}_{T} ; \mathbf{0}, \mathbf{I}\right)$.

If the magnitude $1-\alpha_{t}$ of the noise added at each step is small enough, then the posterior $q\left(x_{t-1} \mid x_{t}\right)$ could be well-approximated by a diagonal Gaussian. Furthermore, if the magnitude $1-\alpha_{1} \ldots \alpha_{T}$ of the total noise added throughout the chain is large enough, then $x_{T}$ could be well-approximated by $\mathcal{N}(0, \mathcal{I})$. These properties suggest learning a model $p_{\theta}\left(x_{t-1} \mid x_{t}\right)$ to approximate the true posterior:
\begin{equation}
p_{\theta}\left(x_{t-1} \mid x_{t}\right):=\mathcal{N}\left(\mu_{\theta}\left(x_{t}\right), \Sigma_{\theta}\left(x_{t}\right)\right),
\end{equation}
which can be used to produce samples $x_{0} \sim p_{\theta}\left(x_{0}\right)$ by starting with Gaussian noise $x_{T} \sim \mathcal{N}(0, \mathcal{I})$ and gradually reducing the noise in a sequence of steps $x_{T-1}, x_{T-2}, \ldots, x_{0}$. To compute this surrogate objective, Ho et al.~\cite{DDPM} found that predicting $\epsilon$ worked best, especially when combined with a reweighted loss function:
\begin{equation}
L_{\text {simple }}:=E_{t, x_{0},\epsilon}\left[\left\|\epsilon-\epsilon_{\theta}\left(x_{t}, t\right)\right\|^{2}\right].
\end{equation}

The DDPM model~\cite{DDPM} shows how to derive $\mu_{\theta}\left(x_{t}\right)$ from $\epsilon_{\theta}\left(x_{t}, t\right)$, and fix $\Sigma_{\theta}$ to a constant. Results from the DDPM show that they can rapidly sample and achieve better log-likelihoods used parameterization and simplified training objective to improve the learning of ${\Sigma _{\theta }}$.

\textbf{Guided Diffusion.} To explicitly incorporate class information into the diffusion process, Dhariwal et al.~\cite{Guided-Diffusion} trained a classifier $f_{\phi}\left(y \mid \mathbf{x}_{t}, t\right)$ on noisy image $x_{t}$ and use gradients $\nabla_{x_{t}} \log p_{\phi}\left(y \mid x_{t}\right)$ to guide the diffusion sampling process toward the target class label $y$. The new resulting perturbed mean $\hat{\mu}_{\theta}\left(x_{t} \mid y\right)$ is given by
\begin{equation}
\hat{\mu}_{\theta}\left(x_{t} \mid y\right)=\mu_{\theta}\left(x_{t} \mid y\right)+s \cdot \Sigma_{\theta}\left(x_{t} \mid y\right) \nabla_{x_{t}} \log p_{\phi}\left(y \mid x_{t}\right),
\end{equation}
where mean $\mu_{\theta}\left(x_{t} \mid y\right)$ and variance $\Sigma_{\theta}\left(x_{t} \mid y\right)$ is perturbed additively by the gradient of the log-probability $\log p_{\phi}\left(y \mid x_{t}\right)$ of a target class $y$ predicted by a classifier. 
The ADM and the one with additional classifier guidance (ADM-G) can achieve results that are better than those of state-of-the-art generative models~\cite{BigGAN}.

\begin{figure*}[t]
    \centering
    \includegraphics[width=\textwidth]{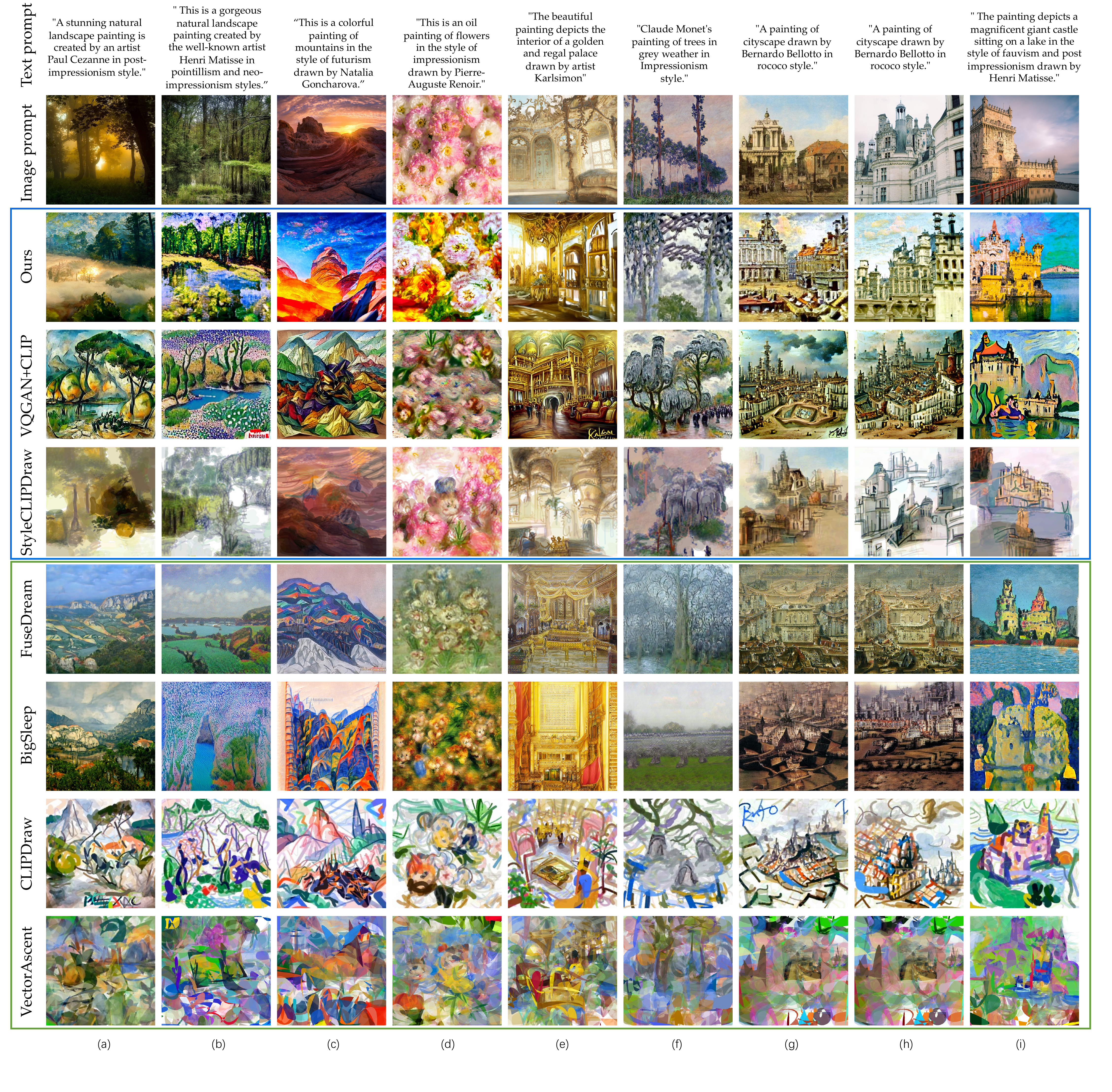}
    \caption{Comparisons of digital art synthesis results using different methods. Methods framed in blue are guided by text and image prompts while methods framed in green are guided by text prompts only.}
    \label{fig:baseline}
\end{figure*}
\begin{figure*}[tb]
    \centering
    \includegraphics[width=\textwidth]{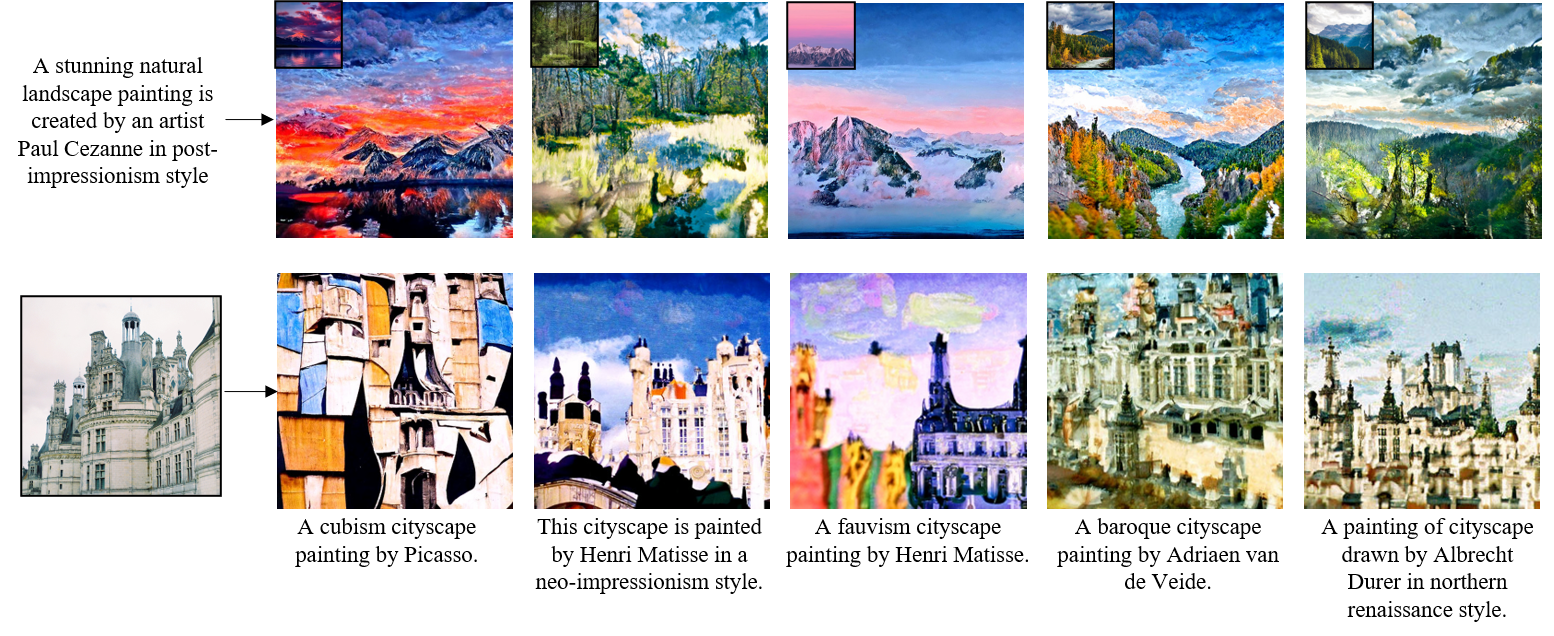}
    \caption{The results are obtained by fixing one mode while the other mode is changed. The first line of the figure fixes the image prompt, and the text prompts are different. The second line fixes the text prompt, and the image prompts are distinct.}
    \label{fig:mgs}
\end{figure*}
\begin{algorithm}[htb]  
  \caption{Multimodal guided diffusion sampling, given \\ a diffusion model $\left(\mu_{\theta}\left(x_{t}\right), \Sigma_{\theta}\left(x_{t}\right)\right)$, guidance function $F_{}\left(x_{t}, t\right)$ and $CLIP$ model}  
  \label{alg1}  
  \begin{algorithmic}[1]  
    \REQUIRE Multimodal guidance $c$, gradient scale $s$, diffusion steps $T$.
		\ENSURE generated image $\hat{x}_{0}$ according to multimodal guidance $c$.
		\STATE $t$ = $T$;
		\STATE $x_{t} \longleftarrow$ sample from $\mathcal{N}\left(0, \mathbf{I}\right)$;
		\REPEAT
		\STATE $t \gets t - 1$ ;
	        \STATE $\mu, \Sigma \leftarrow \mu_{\theta}\left(x_{t}\right), \Sigma_{\theta}\left(x_{t}\right)$;
            \STATE $\hat{\epsilon}_{\theta}\left(x_{t} \mid c\right) \leftarrow \left ( 1-s\right )\cdot {\epsilon _{\theta }}\left (x _{t}\mid \emptyset\right )+s \cdot\epsilon_{\theta}\left(x_{t} \mid c\right)$;
            \STATE $x_{t-1} \leftarrow$ sample from $\mathcal{N}\left(\mu+s\Sigma\nabla_{x_{t}}F_{}\left(x_{t}, t\right)\right)$;
		\UNTIL {$t < 0$} 
  \end{algorithmic}  
\end{algorithm}

\textbf{Classifier-free guidance.} The disadvantage of classifier guidance is that it needs an additional classifier model, thereby complicating the training process. 
Ho and Salimans~\cite{ho2021classifier} presented classifier-free guidance, a methodology for guiding diffusion models that do not demand the training of a separate classifier model. Throughout the training, the tag $y$ in a class-conditional diffusion model $\epsilon_{\theta}\left(x_{t} \mid y\right)$ is substituted with a null tag $\emptyset$ with a defined likelihood for classifier-free guidance.
The output of the model is further extended in the direction of $\epsilon_{\theta}\left(x_{t} \mid y\right)$ and away from $\epsilon_{\theta}\left(x_{t} \mid \emptyset\right)$ during sampling:
\begin{equation}
\hat{\epsilon}_{\theta}\left(x_{t} \mid y\right)=\epsilon_{\theta}\left(x_{t} \mid \emptyset\right)+s \cdot\left(\epsilon_{\theta}\left(x_{t} \mid y\right)-\epsilon_{\theta}\left(x_{t} \mid \emptyset\right)\right).
\end{equation}
The guidance scale is $s \geq 1$. The latent classifier inspired this equation.
\begin{equation}
p^{i}\left(y \mid x_{t}\right) \propto \frac{p\left(x_{t} \mid y\right)}{p\left(x_{t}\right)},
\end{equation}
where gradient is expressed as a function of the true scores $\epsilon^{*}$
\begin{equation}
\begin{aligned}
\nabla_{x_{t}} \log p^{i}\left(x_{t} \mid y\right) & \propto \nabla_{x_{t}} \log p\left(x_{t} \mid y\right)-\nabla_{x_{t}} \log p\left(x_{t}\right), \\
& \propto \epsilon^{*}\left(x_{t} \mid y\right)-\epsilon^{*}\left(x_{t}\right).
\end{aligned}
\end{equation}
As shown in Algorithm~\ref{alg1}, the amended prediction $\hat{\epsilon}$ is then used to steer us towards the multimodal prompts $c$:
\begin{equation}
\hat{\epsilon}_{\theta}\left(x_{t} \mid c\right)=\epsilon_{\theta}\left(x_{t} \mid \emptyset\right)+s \cdot\left(\epsilon_{\theta}\left(x_{t} \mid c\right)-\epsilon_{\theta}\left(x_{t} \mid \emptyset\right)\right).
\end{equation}
Overall, classifier-free guidance has two advantages. First, rather than depending on the information of a separate (and possibly smaller) classification model, it enables a single model to exploit its expertise during guiding. Second, when conditioned on information that is difficult to anticipate using a classifier, it simplifies guiding.

\subsection{CLIP-based Multimodal Guidance}
CLIP~\cite{openai} was presented to acquire visual concepts with natural language supervision and can provide the similarity scores between texts and images.
Several works have used CLIP to steer generative models, such as GANs~\cite{StyleCLIP,StyleGANNADA,CLIPstyler}, toward user-defined text prompts.
In this paper, we leverage a pre-trained CLIP model for text-driven and image-driven art paintings synthesis. 
Both prompts are formulated as the cosine similarity of the created paintings' features to control the semantic content of the generated digital art paintings.

Text prompt $l$ and image prompt $x$ are embedded into the joint embedding space. 
The image encoder $E_{I}$ is time-dependent and trained on noisy pictures. 
$E_{I}^{\prime}$ is the label given to a time-dependent image encoder for noisy pictures. 
The text guidance function can be defined as:
\begin{equation}
F\left(x_{t}, l, t\right)=E_{I}^{\prime}\left(x_{t}, t\right) \cdot E_{L}(l).
\end{equation}
To define the image guiding function, we employ an image encoder finetuned using denoised painting synthesis, similar to how text guidance is generated. The guidance signal at time step $t$ is:
\begin{equation}
F\left(x_{t}, x_{t}^{\prime}, t\right)=E_{I}^{\prime}\left(x_{t}, t\right) \cdot E_{I}^{\prime}\left(x_{t}^{\prime}, t\right).
\end{equation}
The ability to unify image and text guidance simultaneously increases user control flexibility and controllability. 
Both can be readily included in our pipeline (Fig.~\ref{fig:pipeline}).
\begin{equation}
F_{}\left(x_{t}, t\right)=w_{1} F_{}\left(x_{t}, l, t\right)+w_{2} F_{}\left(x_{t}, x_{t}^{\prime}, t\right).
\end{equation}
Users may modify the balance between the two by adjusting each modality's weighting and scale factors.

We replace the classifier with a CLIP model. 
As illustrated in Algorithm~\ref{alg1}, the gradient of the dot product of the multimodal prompts and generated image encoding affect the reverse-process mean: 
\begin{equation}
\hat{\mu}_{\theta}\left(x_{t} \mid c\right)=\mu_{\theta}\left(x_{t} \mid c\right)+s \cdot \Sigma_{\theta}\left(x_{t} \mid c\right) \nabla_{x_{t}}F_{}\left(x_{t}, t\right).
\end{equation}
For MGAD, we employ noised CLIP model, which has been specifically trained to be noise-aware.
\section{EXPERIMENTS}
\subsection{Implementation Details}
For the diffusion model, we used an unconditional model of resolution $512 \times 512$ fine-tuned from OpenAI's class-conditional ImageNet diffusion model~\cite{Guided-Diffusion}. Additionally, we use a secondary diffusion model~\cite{model} pre-trained on Yahoo Flickr Creative Commons 100 Million (YFCC100m)~\cite{yfcc100} dataset to achieve a better performance. For the CLIP model, we used ViT-B/32 released by OpenAI for the Vision Transformer~\cite{vision-transformer}.
The output size of MGAD is $512 \times 512$. 
For sampling, we set $w_{1}$, $w_{2}$ and $s$ to $1.0$, $1.0$ and $5,000$, respectively. 
We utilize the U-Net~\cite{UNet} architecture based on Wide-ResNet~\cite{wideres}. 
The model has seven downsampling and seven upsampling layers. The $4 \times 4$ feature is generated from the $512 \times 512$ input image via one input convolution and fine Resblocks. From the $32 \times 32$ to $4 \times 4$ resolution, self-attention blocks are added to the Resblocks. 
The clip model for generating is ViT-B/32. 
To ensure the quality of the results and to maintain the consistency of the parameters, the diffusion step and the time step used for the experiments in this work are set to $2,000$. 
Remarkable results are generated when the time step is set to $500$ or more. 
One upsampling step takes approximately $63$ milliseconds on a single A40 GPU.
For more hyperparameter settings, please refer to the Supplementary Material.

\begin{table*}[htbp]
  \caption{Quantitative comparisons for digital art generation. We compute the average learned perceptual image patch similarity (LPIPS) and difference Hash (dHash) values of results to measure diversity. The best results are presented in \textbf{bold}.}
  \label{tab:quantity}
  \centering
  \begin{tabular}{cccccccc}
    \toprule
         & Ours & Vector Ascent  & FuseDream & BigSleep & CLIPDraw & StyleCLIPDraw & VQGAN+CLIP \\
    \midrule
    LPIPS ↓ & \textbf{0.458} & 0.614 & 0.58 & 0.544 & 0.735 & 0.489 & 0.618 \\
    \midrule
    dHash ↑ & \textbf{0.713} & 0.551 & 0.616 & 0.502 & 0.534 & 0.519 & 0.516 \\
    \bottomrule
  \end{tabular}
  

\end{table*}
\begin{table*}[htbp]
  \caption{User study results. Each number represents the percentage of votes that the results generated by the other models received when selected against our results.}
  \label{tab:userstudy}
  \centering
  \begin{tabular}{ccccccc}
    \toprule
          & VectorAscent & FuseDream & BigSleep & CLIPDraw & StyleCLIPDraw & VQGAN+CLIP \\
    \midrule
    Preference rate & 10.9\% & 29.3\% & 27.5\% & 9.5\% & 29.4\% & 26.3\% \\
    \bottomrule
  \end{tabular}
\end{table*}

\subsection{Qualitative Evaluation}

\paragraph{Baselines} To demonstrate the performance of our method, we compare MGAD with SOTA digital painting synthesis works, including VQGAN-CLIP~\cite{vqgan-clip}, FuseDream~\cite{fusedream}, BigSleep~\cite{bigsleep}, StyleCLIPDraw~\cite{styleclipdraw}, CLIPDraw~\cite{clipdraw}, and VectorAscent~\cite{vector}. 
Figure~\ref{fig:baseline} shows the digital art generation results.
VQGAN-CLIP~\cite{esser2021taming} and StyleCLIPDraw~\cite{styleclipdraw} use two modal prompts. 
The others are only able to use text prompts. 
VQGAN-CLIP~\cite{esser2021taming}, FuseDream~\cite{fusedream}, and BIGGAN-CLIP~\cite{bigsleep} utilize CLIP-loss between the prompts and the generated results to realize latent code optimization. Still, problems are encountered in generating paintings with similar brush strokes and structures. 
Their results have similar repeated textures in different image locations. 
VQGAN-CLIP~\cite{esser2021taming} can make use of the two modal prompts. However, it may not mimic the painter’s strokes well (e.g., the 4$^{th}$ row in Figures~\ref{fig:baseline}a and \ref{fig:baseline}d-\ref{fig:baseline}i) during the generation process, and the generated results do not resemble the image prompt very much. 
FuseDream and BigSleep use pre-trained BigGANs~\cite{BigGAN} and CLIP to achieve high-resolution text-to-image generation. 
The weights of the generator are frozen; only the latent $Z$ vectors are optimized, so they often generate repetitive patterns in the results (e.g., the 6$^{th}$ and 7$^{th}$ rows in Figures~\ref{fig:baseline}g-\ref{fig:baseline}h). 
CLIPDraw~\cite{clipdraw} and  VectorAscent~\cite{vector} rely on CLIP because of its aligned text and image encoders and diffvg\cite{diffvg}, which is a differentiable vector graphics rasterizer used to generate raster images from vector paths. 
By using gradient ascent, they can optimize for a vector graphic whose rasterization has high similarity with a user-provided caption, backpropagating through CLIP~\cite{openai} and diffvg~\cite{diffvg} to the vector graphic parameters. 
Their results have resemblances to the text prompt. 
However, this type of method can only produce discontinuous outlines that resemble the touch of a watercolor brush or blocks of color, unlike human paintings (e.g., the 5$^{th}$, 8$^{th}$ and 9$^{th}$ rows in Figure~\ref{fig:baseline}). 
The StyleCLIPDraw~\cite{styleclipdraw} adds a stylized module to the CLIPDraw~\cite{clipdraw}, thereby making the stroke styles more diverse, but it still falls short in approaching the realism of the content described by the text prompts (e.g., the 5$^{th}$ and 8$^{th}$ rows in Figure~\ref{fig:baseline}).

Compared with baselines, the proposed MGAD can combine the semantic content of visual and language modalities to generate the requested digital artworks. CLIP~\cite{openai} provides a good understanding of the different modal prompts at the level of semantic content. 
Diffusion models~\cite{Guided-Diffusion} can balance high quality and diversity in the generation of images. Therefore, using the MAGD, generating complete and sophisticated artworks with skills that meet the user's requirements for the content and the texture and style that closely resembles the paintings created by the mentioned artist is possible.

\begin{figure*}[bt]
    \centering
    \includegraphics[width=\textwidth]{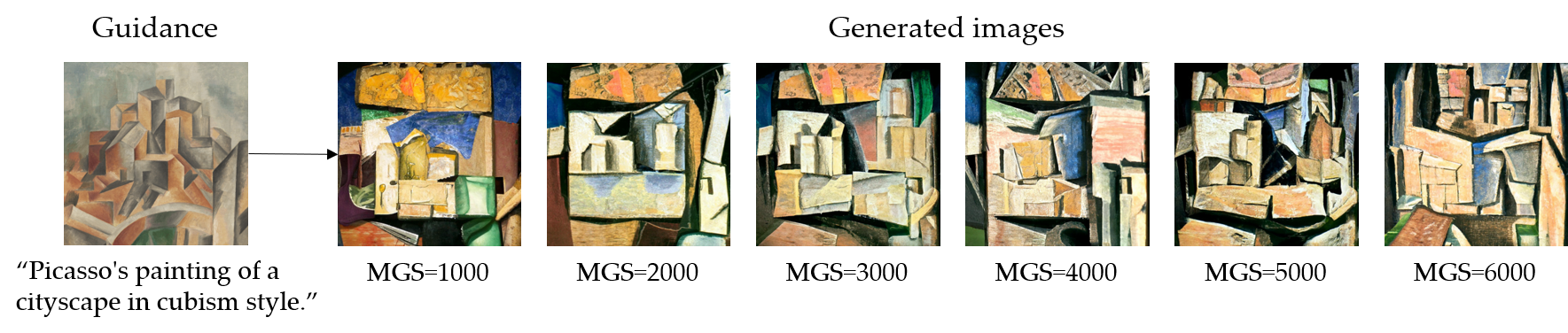}
    \vspace{-0.6cm} 
    \caption{Effect of multimodal guidance scale (MGS) on the generated results.}
    \vspace{-0.4cm} 
    \label{fig:scale}
\end{figure*}

\paragraph{Strong Understanding for Wide Prompts} By observing the content of the prompts in Figure~\ref{fig:teaser}, we find that MGAD can meet the prompts that present a variety of requirements. There is a good understanding of the painter, painting style, color, and texture, and can show their differences. Image prompt can generate good results regardless of whether it is a real image or a painting. MGAD's ability to blend the two modalities and find commonalities between them is outstanding.

\begin{figure*}[tb]
    \centering
    \includegraphics[width=\textwidth]{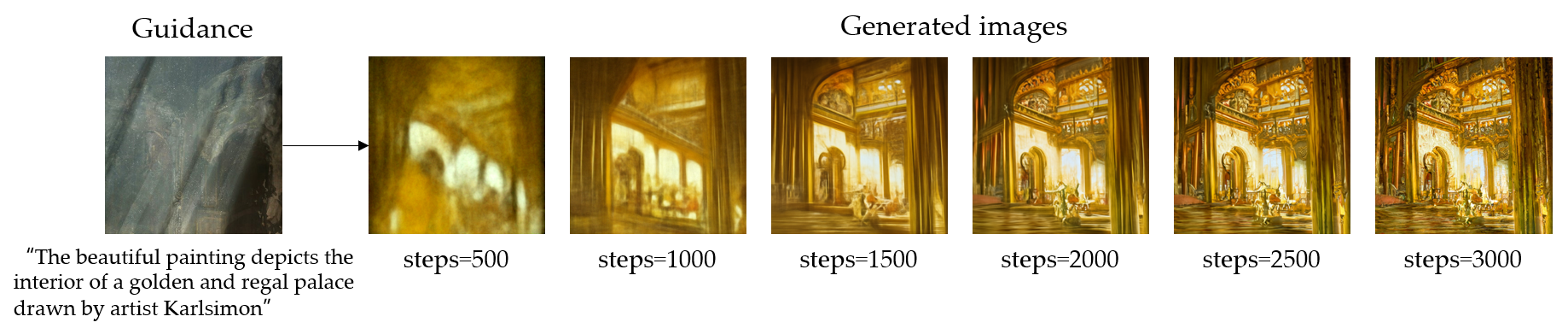}
    \vspace{-0.6cm} 
    \caption{Results of different diffusion steps.}
    \vspace{-0.4cm} 
    \label{fig:diffusionstep}
\end{figure*}
\begin{figure*}[tb]
    \centering
    \includegraphics[width=\textwidth]{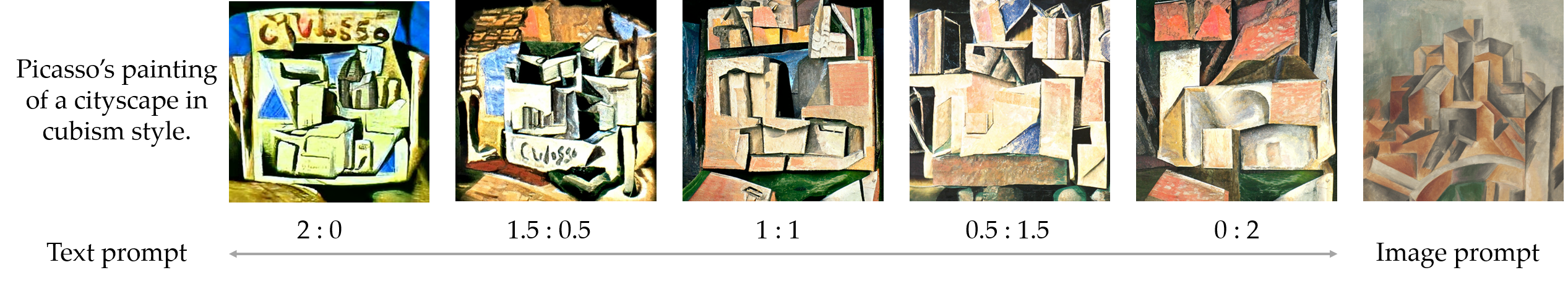}
    \vspace{-0.6cm} 
    \caption{Visualizations of the trade-off between a text prompt and an image prompt.}
    \vspace{-0.4cm} 
    \label{fig:text-image}
\end{figure*}

\subsection{Quantitative Evaluation}

To demonstrate the advantages of our method in terms of the quality and diversity of digital art synthesis, we compare it with the existing methods, namely, VectorAscent~\cite{vector}, FuseDream~\cite{fusedream}, BigSleep~\cite{bigsleep},  CLIPDraw~\cite{clipdraw}, StyleCLIPDraw~\cite{styleclipdraw} and  VQGAN-CLIP~\cite{vqgan-clip} using LPIPS~\cite{LPIPS}, and dHash~\cite{dhash}. User study is also conducted.

\paragraph{LPIPS for diversity evaluation} 
LPIPS~\cite{LPIPS} measures the perceptual similarity between two images. We calculate the LPIPS score between paired images generated from the same prompts for guidance (Table~\ref{tab:quantity}). 
Intuitively, the lower the LPIPS score is, the lesser the similarity of the generated images and the higher the diversity of the generated paintings will be. 
The results show that our model outperforms baseline models by showing the best LPIPS score and surpassing others by a margin. This finding indicates that the LPIPS score results have almost the same tendencies as the score of the user study.


\paragraph{dHash} 
dHash~\cite{dhash} is a hash mapping method based on pixel point level to measure the similarity between images. As an implementation, dHash is nearly identical to aHash, but it performs much better. A value of 0 indicates the same hash and likely a similar picture. The comparison between the different model dHash values is shown in the bottom row of Table~\ref{tab:quantity}. Our model scores significantly superior to other models on this evaluation metric. This result demonstrates that MGAD is notable in terms of diversity.
 
\paragraph{User Study}
We conduct a user study to further compare our method. The user study was divided into two groups.
The first group compares the models using text as guidance, and the second group uses text and image as guidance. VectorAscent~\cite{vector}, FuseDream~\cite{fusedream}, BigSleep~\cite{bigsleep}, and CLIPDraw~\cite{clipdraw} are in the first group, and the other models are in the second group. 
We invited $82$ users to participate in the first study and invited $89$ users to participate in the second study to evaluate the results of different approaches. 

Given the guidance, for each set, we show the result generated by our approach and the output from another randomly selected method for comparison and ask the user to select which digital artwork has better effects. Participants in the first group wrote $27$ questions and collected $2,214$ votes. Participants in the second group reported $15$ questions and gathered $1,335$ votes. We calculated the percentage of votes where the existing methods are superior to ours and show the statistical results in Table~\ref{tab:userstudy}. The results show that most people prefer our approach compared with other methods.

\subsection{Ablation Study}

\paragraph{The Influence of Multimodal Guidance Scale} The MGS decides how strongly the result should match the multimodal prompts. We compared the generated results for different MGS values to verify the impact of multimodal guidance. As shown in Figure 6, the generated results are more creative and uncertain when using a lower MGS. Meanwhile, using a higher MGS brings the generated results closer to the semantic content of the prompts and the user requirements. With the appropriate scale of multimodal guidance, the results can be made to fit the prompt content while exploiting the creativity of the model. Users can adjust the scaling factor to control how diverse they expect the generated images to be.

\paragraph{Impact of Diffusion Steps} 
To investigate the effect of the number of diffusion steps on the quality and generation speed of the generated images, we tested the generation at different diffusion steps. The experiments show that the sampling process is faster when the number of diffusion steps is small, although the generated images are unclear. After the diffusion steps gradually increase, the image quality improves, and the sampling time increases. However, as shown in Fig.~\ref{fig:diffusionstep}, after a certain number of diffusion steps, the image quality no longer changes significantly. In the trade-off between generation time and generation quality, we usually set the number of diffusion steps to 2000.

\paragraph{Balance between Text and Image Prompts}
As shown in Figure~\ref{fig:text-image}, we can adjust the degree of guidance of both for the final generated results by setting the values of the weight $w_1$ for the image prompt and the weight $w_2$ for the text prompt. During our experiments, we found that if the difference between image prompt and text prompt in semantic content is too large, then it will lead to the generation of unsatisfactory results. For this reason, we propose the adaptive prompt discarding module, which calculates the similarity between image prompt and text prompt and then compares the weights of the two to decide whether one of the modalities should be discarded to obtain a better generation result. 

\section{CONCLUSIONS AND FUTURE WORK}
In this paper, we propose the MGAD model, which is a digital artwork generation model that combines multimodal prompt guidance with classifier-free diffusion model guidance. To enable MGAD to control the generation of digital artwork with the semantic content of multimodal prompts, we propose multimodal guidance loss to constrain the diversity of models and the content similarity with prompts. Adequate experiments show that our approach can achieve a balance between the quality and diversity of digital artwork generation again. In future work, we will focus more on the disentangling and fusion of multimodal prompts and involving art stroke/model-based techniques to make the model more general, creative, and user friendly.

\begin{acks}
This work was supported by National Key R\&D Program of China under No. 2020AAA0106200, by National Natural Science Foundation of China under Nos. 61832016, U20B2070, 6210070958, 62102162, and by Open Projects Program of NLPR.
\end{acks}

\bibliographystyle{ACM-Reference-Format}
\balance
\bibliography{MultiDiffusion}


\begin{thebibliography}{78}


\ifx \showCODEN    \undefined \def \showCODEN     #1{\unskip}     \fi
\ifx \showDOI      \undefined \def \showDOI       #1{#1}\fi
\ifx \showISBNx    \undefined \def \showISBNx     #1{\unskip}     \fi
\ifx \showISBNxiii \undefined \def \showISBNxiii  #1{\unskip}     \fi
\ifx \showISSN     \undefined \def \showISSN      #1{\unskip}     \fi
\ifx \showLCCN     \undefined \def \showLCCN      #1{\unskip}     \fi
\ifx \shownote     \undefined \def \shownote      #1{#1}          \fi
\ifx \showarticletitle \undefined \def \showarticletitle #1{#1}   \fi
\ifx \showURL      \undefined \def \showURL       {\relax}        \fi
\providecommand\bibfield[2]{#2}
\providecommand\bibinfo[2]{#2}
\providecommand\natexlab[1]{#1}
\providecommand\showeprint[2][]{arXiv:#2}

\bibitem[yfc(2022)]%
        {yfcc100}
 \bibinfo{year}{2022}\natexlab{}.
\newblock \bibinfo{booktitle}{\emph{Yahoo Flickr Creative Commons 100 Million
  (YFCC100m) dataset}}.
\newblock
\urldef\tempurl%
\url{http://projects.dfki.uni-kl.de/yfcc100m/}
\showURL{%
\tempurl}


\bibitem[Adverb(2022)]%
        {bigsleep}
\bibfield{author}{\bibinfo{person}{Adverb}.} \bibinfo{year}{2022}\natexlab{}.
\newblock \bibinfo{booktitle}{\emph{The BigSleep: BigGAN+CLIP}}.
\newblock
\urldef\tempurl%
\url{https://colab.research.google.com/drive/1NCceX2mbiKOSlAd_o7IU7nA9UskKN5WR?usp=sharing#scrollTo=WtlDVVMvzMUd}
\showURL{%
\tempurl}


\bibitem[Alvarez et~al\mbox{.}(2021)]%
        {Alvarez:2021:IDR}
\bibfield{author}{\bibinfo{person}{Luis Alvarez}, \bibinfo{person}{Nelson
  Monzón}, {and} \bibinfo{person}{Jean-Michel Morel}.}
  \bibinfo{year}{2021}\natexlab{}.
\newblock \showarticletitle{{Interactive Design of Random Aesthetic Abstract
  Textures by Composition Principles}}.
\newblock \bibinfo{journal}{\emph{Leonardo}} \bibinfo{volume}{54},
  \bibinfo{number}{2} (\bibinfo{year}{2021}), \bibinfo{pages}{179--184}.
\newblock


\bibitem[Brock et~al\mbox{.}(2018)]%
        {BigGAN}
\bibfield{author}{\bibinfo{person}{Andrew Brock}, \bibinfo{person}{Jeff
  Donahue}, {and} \bibinfo{person}{Karen Simonyan}.}
  \bibinfo{year}{2018}\natexlab{}.
\newblock \showarticletitle{Large Scale GAN Training for High Fidelity Natural
  Image Synthesis}. In \bibinfo{booktitle}{\emph{International Conference on
  Learning Representations (ICLR)}}.
\newblock


\bibitem[Buchner(2021)]%
        {dhash}
\bibfield{author}{\bibinfo{person}{Johannes Buchner}.}
  \bibinfo{year}{2021}\natexlab{}.
\newblock \bibinfo{booktitle}{\emph{ImageHash:An image hashing library written
  in Python.}}
\newblock
\urldef\tempurl%
\url{https://github.com/JohannesBuchner/imagehash}
\showURL{%
\tempurl}


\bibitem[Chen et~al\mbox{.}(2021a)]%
        {chen2021dualast}
\bibfield{author}{\bibinfo{person}{Haibo Chen}, \bibinfo{person}{Lei Zhao},
  \bibinfo{person}{Zhizhong Wang}, \bibinfo{person}{Huiming Zhang},
  \bibinfo{person}{Zhiwen Zuo}, \bibinfo{person}{Ailin Li},
  \bibinfo{person}{Wei Xing}, {and} \bibinfo{person}{Dongming Lu}.}
  \bibinfo{year}{2021}\natexlab{a}.
\newblock \showarticletitle{{DualAST}: Dual Style-Learning Networks for
  Artistic Style Transfer}. In \bibinfo{booktitle}{\emph{IEEE/CVF Conference on
  Computer Vision and Pattern Recognition (CVPR)}}. \bibinfo{pages}{872--881}.
\newblock


\bibitem[Chen et~al\mbox{.}(2021b)]%
        {Chen:2021:DIS}
\bibfield{author}{\bibinfo{person}{Haibo Chen}, \bibinfo{person}{Lei Zhao},
  \bibinfo{person}{Huiming Zhang}, \bibinfo{person}{Zhizhong Wang},
  \bibinfo{person}{Zhiwen Zuo}, \bibinfo{person}{Ailin Li},
  \bibinfo{person}{Wei Xing}, {and} \bibinfo{person}{Dongming Lu}.}
  \bibinfo{year}{2021}\natexlab{b}.
\newblock \showarticletitle{Diverse Image Style Transfer via Invertible
  Cross-Space Mapping}. In \bibinfo{booktitle}{\emph{IEEE/CVF International
  Conference on Computer Vision (ICCV)}}. \bibinfo{pages}{14860--14869}.
\newblock


\bibitem[Chen et~al\mbox{.}(2020)]%
        {chen2020uniter}
\bibfield{author}{\bibinfo{person}{Yen-Chun Chen}, \bibinfo{person}{Linjie Li},
  \bibinfo{person}{Licheng Yu}, \bibinfo{person}{Ahmed El~Kholy},
  \bibinfo{person}{Faisal Ahmed}, \bibinfo{person}{Zhe Gan},
  \bibinfo{person}{Yu Cheng}, {and} \bibinfo{person}{Jingjing Liu}.}
  \bibinfo{year}{2020}\natexlab{}.
\newblock \showarticletitle{Uniter: Universal image-text representation
  learning}. In \bibinfo{booktitle}{\emph{European Conference on Computer
  Vision (ECCV)}}. \bibinfo{pages}{104--120}.
\newblock


\bibitem[Choi et~al\mbox{.}(2021)]%
        {ILVR}
\bibfield{author}{\bibinfo{person}{Jooyoung Choi}, \bibinfo{person}{Sungwon
  Kim}, \bibinfo{person}{Yonghyun Jeong}, \bibinfo{person}{Youngjune Gwon},
  {and} \bibinfo{person}{Sungroh Yoon}.} \bibinfo{year}{2021}\natexlab{}.
\newblock \showarticletitle{ILVR: Conditioning Method for Denoising Diffusion
  Probabilistic Models}. In \bibinfo{booktitle}{\emph{IEEE/CVF International
  Conference on Computer Vision (ICCV)}}. \bibinfo{pages}{14347--14356}.
\newblock


\bibitem[Crowson and AI(2022)]%
        {model}
\bibfield{author}{\bibinfo{person}{Katherine Crowson} {and}
  \bibinfo{person}{Chainbreakers AI}.} \bibinfo{year}{2022}\natexlab{}.
\newblock \bibinfo{booktitle}{\emph{Diffusion 512x512, secondary model
  method.}}
\newblock
\urldef\tempurl%
\url{https://github.com/crowsonkb/v-diffusion-pytorch}
\showURL{%
\tempurl}


\bibitem[Deng et~al\mbox{.}(2021)]%
        {deng2021arbitrary}
\bibfield{author}{\bibinfo{person}{Yingying Deng}, \bibinfo{person}{Fan Tang},
  \bibinfo{person}{Weiming Dong}, \bibinfo{person}{Haibin Huang},
  \bibinfo{person}{Chongyang Ma}, {and} \bibinfo{person}{Changsheng Xu}.}
  \bibinfo{year}{2021}\natexlab{}.
\newblock \showarticletitle{Arbitrary video style transfer via multi-channel
  correlation}. In \bibinfo{booktitle}{\emph{Proceedings of the AAAI Conference
  on Artificial Intelligence}}, Vol.~\bibinfo{volume}{35}.
  \bibinfo{pages}{1210--1217}.
\newblock


\bibitem[Deng et~al\mbox{.}(2020a)]%
        {deng2020exploring}
\bibfield{author}{\bibinfo{person}{Yingying Deng}, \bibinfo{person}{Fan Tang},
  \bibinfo{person}{Weiming Dong}, \bibinfo{person}{Chongyang Ma},
  \bibinfo{person}{Feiyue Huang}, \bibinfo{person}{Oliver Deussen}, {and}
  \bibinfo{person}{Changsheng Xu}.} \bibinfo{year}{2020}\natexlab{a}.
\newblock \showarticletitle{Exploring the representativity of art paintings}.
\newblock \bibinfo{journal}{\emph{IEEE Transactions on Multimedia}}
  \bibinfo{volume}{23} (\bibinfo{year}{2020}), \bibinfo{pages}{2794--2805}.
\newblock


\bibitem[Deng et~al\mbox{.}(2022)]%
        {deng2021stytr2}
\bibfield{author}{\bibinfo{person}{Yingying Deng}, \bibinfo{person}{Fan Tang},
  \bibinfo{person}{Weiming Dong}, \bibinfo{person}{Chongyang Ma},
  \bibinfo{person}{Xingjia Pan}, \bibinfo{person}{Lei Wang}, {and}
  \bibinfo{person}{Changsheng Xu}.} \bibinfo{year}{2022}\natexlab{}.
\newblock \showarticletitle{StyTr$^2$: Image Style Transfer with Transformers}.
  In \bibinfo{booktitle}{\emph{IEEE/CVF Conference on Computer Vision and
  Pattern Recognition (CVPR)}}.
\newblock


\bibitem[Deng et~al\mbox{.}(2020b)]%
        {deng2020arbitrary}
\bibfield{author}{\bibinfo{person}{Yingying Deng}, \bibinfo{person}{Fan Tang},
  \bibinfo{person}{Weiming Dong}, \bibinfo{person}{Wen Sun},
  \bibinfo{person}{Feiyue Huang}, {and} \bibinfo{person}{Changsheng Xu}.}
  \bibinfo{year}{2020}\natexlab{b}.
\newblock \showarticletitle{Arbitrary style transfer via multi-adaptation
  network}. In \bibinfo{booktitle}{\emph{Proceedings of the 28th ACM
  International Conference on Multimedia}}. \bibinfo{pages}{2719--2727}.
\newblock


\bibitem[Deng et~al\mbox{.}(2019)]%
        {deng2019selective}
\bibfield{author}{\bibinfo{person}{Yingying Deng}, \bibinfo{person}{Fan Tang},
  \bibinfo{person}{Weiming Dong}, \bibinfo{person}{Fuzhang Wu},
  \bibinfo{person}{Oliver Deussen}, {and} \bibinfo{person}{Changsheng Xu}.}
  \bibinfo{year}{2019}\natexlab{}.
\newblock \showarticletitle{Selective clustering for representative paintings
  selection}.
\newblock \bibinfo{journal}{\emph{Multimedia Tools and Applications}}
  \bibinfo{volume}{78}, \bibinfo{number}{14} (\bibinfo{year}{2019}),
  \bibinfo{pages}{19305--19323}.
\newblock


\bibitem[Desai and Johnson(2021)]%
        {desai2021virtex}
\bibfield{author}{\bibinfo{person}{Karan Desai} {and} \bibinfo{person}{Justin
  Johnson}.} \bibinfo{year}{2021}\natexlab{}.
\newblock \showarticletitle{Virtex: Learning visual representations from
  textual annotations}. In \bibinfo{booktitle}{\emph{IEEE/CVF Conference on
  Computer Vision and Pattern Recognition (CVPR)}}.
  \bibinfo{pages}{11162--11173}.
\newblock


\bibitem[Dhariwal and Nichol(2021)]%
        {Guided-Diffusion}
\bibfield{author}{\bibinfo{person}{Prafulla Dhariwal} {and}
  \bibinfo{person}{Alex Nichol}.} \bibinfo{year}{2021}\natexlab{}.
\newblock \showarticletitle{Diffusion Models Beat GANs on Image Synthesis}. In
  \bibinfo{booktitle}{\emph{Advances in Neural Information Processing Systems
  (NeurIPS)}}.
\newblock


\bibitem[Dosovitskiy et~al\mbox{.}(2020)]%
        {vision-transformer}
\bibfield{author}{\bibinfo{person}{Alexey Dosovitskiy}, \bibinfo{person}{Lucas
  Beyer}, \bibinfo{person}{Alexander Kolesnikov}, \bibinfo{person}{Dirk
  Weissenborn}, \bibinfo{person}{Xiaohua Zhai}, \bibinfo{person}{Thomas
  Unterthiner}, \bibinfo{person}{Mostafa Dehghani}, \bibinfo{person}{Matthias
  Minderer}, \bibinfo{person}{Georg Heigold}, \bibinfo{person}{Sylvain Gelly},
  {et~al\mbox{.}}} \bibinfo{year}{2020}\natexlab{}.
\newblock \showarticletitle{An image is worth 16x16 words: Transformers for
  image recognition at scale}. In \bibinfo{booktitle}{\emph{International
  Conference on Learning Representations (ICLR)}}.
\newblock


\bibitem[Esser et~al\mbox{.}(2021)]%
        {esser2021taming}
\bibfield{author}{\bibinfo{person}{Patrick Esser}, \bibinfo{person}{Robin
  Rombach}, {and} \bibinfo{person}{Bjorn Ommer}.}
  \bibinfo{year}{2021}\natexlab{}.
\newblock \showarticletitle{Taming transformers for high-resolution image
  synthesis}. In \bibinfo{booktitle}{\emph{IEEE/CVF Conference on Computer
  Vision and Pattern Recognition (CVPR)}}. \bibinfo{pages}{12873--12883}.
\newblock


\bibitem[Frans et~al\mbox{.}(2021)]%
        {clipdraw}
\bibfield{author}{\bibinfo{person}{Kevin Frans}, \bibinfo{person}{LB Soros},
  {and} \bibinfo{person}{Olaf Witkowski}.} \bibinfo{year}{2021}\natexlab{}.
\newblock \showarticletitle{{CLIPDraw}: Exploring Text-to-Drawing Synthesis
  through Language-Image Encoders}.
\newblock \bibinfo{journal}{\emph{arXiv preprint arXiv:2106.14843}}
  (\bibinfo{year}{2021}).
\newblock


\bibitem[Gal et~al\mbox{.}(2021)]%
        {StyleGANNADA}
\bibfield{author}{\bibinfo{person}{Rinon Gal}, \bibinfo{person}{Or Patashnik},
  \bibinfo{person}{Haggai Maron}, \bibinfo{person}{Gal Chechik}, {and}
  \bibinfo{person}{Daniel Cohen-Or}.} \bibinfo{year}{2021}\natexlab{}.
\newblock \bibinfo{title}{StyleGAN-NADA: CLIP-Guided Domain Adaptation of Image
  Generators}.
\newblock
\newblock
\showeprint[arxiv]{2108.00946}~[cs.CV]


\bibitem[Goodfellow et~al\mbox{.}(2014)]%
        {GAN}
\bibfield{author}{\bibinfo{person}{Ian Goodfellow}, \bibinfo{person}{Jean
  Pouget-Abadie}, \bibinfo{person}{Mehdi Mirza}, \bibinfo{person}{Bing Xu},
  \bibinfo{person}{David Warde-Farley}, \bibinfo{person}{Sherjil Ozair},
  \bibinfo{person}{Aaron Courville}, {and} \bibinfo{person}{Yoshua Bengio}.}
  \bibinfo{year}{2014}\natexlab{}.
\newblock \showarticletitle{Generative Adversarial Nets}. In
  \bibinfo{booktitle}{\emph{Neural Information Processing Systems (NIPS)}}.
\newblock


\bibitem[Gu et~al\mbox{.}(2021)]%
        {VQ-Diffusion}
\bibfield{author}{\bibinfo{person}{Shuyang Gu}, \bibinfo{person}{Dong Chen},
  \bibinfo{person}{Jianmin Bao}, \bibinfo{person}{Fang Wen},
  \bibinfo{person}{Bo Zhang}, \bibinfo{person}{Dongdong Chen},
  \bibinfo{person}{Lu Yuan}, {and} \bibinfo{person}{Baining Guo}.}
  \bibinfo{year}{2021}\natexlab{}.
\newblock \showarticletitle{Vector Quantized Diffusion Model for Text-to-Image
  Synthesis}.
\newblock \bibinfo{journal}{\emph{arXiv preprint arXiv:2111.14822}}
  (\bibinfo{year}{2021}).
\newblock


\bibitem[Hall et~al\mbox{.}(2015)]%
        {hall2015cross}
\bibfield{author}{\bibinfo{person}{Peter Hall}, \bibinfo{person}{Hongping Cai},
  \bibinfo{person}{Qi Wu}, {and} \bibinfo{person}{Tadeo Corradi}.}
  \bibinfo{year}{2015}\natexlab{}.
\newblock \showarticletitle{Cross-depiction problem: Recognition and synthesis
  of photographs and artwork}.
\newblock \bibinfo{journal}{\emph{Computational Visual Media}}
  \bibinfo{volume}{1}, \bibinfo{number}{2} (\bibinfo{year}{2015}),
  \bibinfo{pages}{91--103}.
\newblock


\bibitem[Ho et~al\mbox{.}(2020)]%
        {DDPM}
\bibfield{author}{\bibinfo{person}{Jonathan Ho}, \bibinfo{person}{Ajay Jain},
  {and} \bibinfo{person}{Pieter Abbeel}.} \bibinfo{year}{2020}\natexlab{}.
\newblock \showarticletitle{Denoising Diffusion Probabilistic Models}.
\newblock \bibinfo{journal}{\emph{arXiv: Learning}} (\bibinfo{year}{2020}).
\newblock


\bibitem[Ho and Salimans(2021)]%
        {ho2021classifier}
\bibfield{author}{\bibinfo{person}{Jonathan Ho} {and} \bibinfo{person}{Tim
  Salimans}.} \bibinfo{year}{2021}\natexlab{}.
\newblock \showarticletitle{Classifier-Free Diffusion Guidance}. In
  \bibinfo{booktitle}{\emph{NeurIPS 2021 Workshop on Deep Generative Models and
  Downstream Applications}}.
\newblock


\bibitem[Huang and Belongie(2017)]%
        {Huang:2017:AdaIn}
\bibfield{author}{\bibinfo{person}{Xun Huang} {and} \bibinfo{person}{Serge
  Belongie}.} \bibinfo{year}{2017}\natexlab{}.
\newblock \showarticletitle{Arbitrary style transfer in real-time with adaptive
  instance normalization}. In \bibinfo{booktitle}{\emph{IEEE International
  Conference on Computer Vision (ICCV)}}. \bibinfo{publisher}{IEEE},
  \bibinfo{pages}{1501--1510}.
\newblock


\bibitem[Huang et~al\mbox{.}(2022)]%
        {huang2022dualface}
\bibfield{author}{\bibinfo{person}{Zhengyu Huang}, \bibinfo{person}{Yichen
  Peng}, \bibinfo{person}{Tomohiro Hibino}, \bibinfo{person}{Chunqi Zhao},
  \bibinfo{person}{Haoran Xie}, \bibinfo{person}{Tsukasa Fukusato}, {and}
  \bibinfo{person}{Kazunori Miyata}.} \bibinfo{year}{2022}\natexlab{}.
\newblock \showarticletitle{dualface: Two-stage drawing guidance for freehand
  portrait sketching}.
\newblock \bibinfo{journal}{\emph{Computational Visual Media}}
  \bibinfo{volume}{8}, \bibinfo{number}{1} (\bibinfo{year}{2022}),
  \bibinfo{pages}{63--77}.
\newblock


\bibitem[Huo and Yoon(2021)]%
        {huo2021survey}
\bibfield{author}{\bibinfo{person}{Yuchi Huo} {and} \bibinfo{person}{Sung-eui
  Yoon}.} \bibinfo{year}{2021}\natexlab{}.
\newblock \showarticletitle{A survey on deep learning-based Monte Carlo
  denoising}.
\newblock \bibinfo{journal}{\emph{Computational Visual Media}}
  \bibinfo{volume}{7}, \bibinfo{number}{2} (\bibinfo{year}{2021}),
  \bibinfo{pages}{169--185}.
\newblock


\bibitem[Jain(2021)]%
        {vector}
\bibfield{author}{\bibinfo{person}{Ajay Jain}.}
  \bibinfo{year}{2021}\natexlab{}.
\newblock \bibinfo{booktitle}{\emph{VectorAscent: Generate vector graphics from
  a textual description}}.
\newblock
\urldef\tempurl%
\url{https://github.com/ajayjain/VectorAscent}
\showURL{%
\tempurl}


\bibitem[Johnson et~al\mbox{.}(2016)]%
        {johnson2016perceptual}
\bibfield{author}{\bibinfo{person}{Justin Johnson}, \bibinfo{person}{Alexandre
  Alahi}, {and} \bibinfo{person}{Li Fei-Fei}.} \bibinfo{year}{2016}\natexlab{}.
\newblock \showarticletitle{Perceptual losses for real-time style transfer and
  super-resolution}. In \bibinfo{booktitle}{\emph{European Conference on
  Computer Vision (ECCV)}}. \bibinfo{pages}{694--711}.
\newblock


\bibitem[Jolicoeur-Martineau et~al\mbox{.}(2021)]%
        {AlexiaJolicoeurMartineau2021AdversarialSM}
\bibfield{author}{\bibinfo{person}{Alexia Jolicoeur-Martineau},
  \bibinfo{person}{R{\'e}mi Pich{\'e}-Taillefer}, \bibinfo{person}{Ioannis
  Mitliagkas}, {and} \bibinfo{person}{Remi~Tachet des Combes}.}
  \bibinfo{year}{2021}\natexlab{}.
\newblock \showarticletitle{Adversarial score matching and improved sampling
  for image generation}. In \bibinfo{booktitle}{\emph{International Conference
  on Learning Representations (ICLR)}}.
\newblock


\bibitem[Karras et~al\mbox{.}(2018)]%
        {StyleGAN}
\bibfield{author}{\bibinfo{person}{Tero Karras}, \bibinfo{person}{Samuli
  Laine}, {and} \bibinfo{person}{Timo Aila}.} \bibinfo{year}{2018}\natexlab{}.
\newblock \showarticletitle{A Style-Based Generator Architecture for Generative
  Adversarial Networks}.
\newblock \bibinfo{journal}{\emph{arXiv: Neural and Evolutionary Computing}}
  (\bibinfo{year}{2018}).
\newblock


\bibitem[Karras et~al\mbox{.}(2020)]%
        {StyleGAN2}
\bibfield{author}{\bibinfo{person}{Tero Karras}, \bibinfo{person}{Samuli
  Laine}, \bibinfo{person}{Miika Aittala}, \bibinfo{person}{Janne Hellsten},
  \bibinfo{person}{Jaakko Lehtinen}, {and} \bibinfo{person}{Timo Aila}.}
  \bibinfo{year}{2020}\natexlab{}.
\newblock \showarticletitle{Analyzing and Improving the Image Quality of
  {StyleGAN}}. In \bibinfo{booktitle}{\emph{IEEE/CVF Conference on Computer
  Vision and Pattern Recognition (CVPR)}}. \bibinfo{pages}{8107--8116}.
\newblock


\bibitem[Kim et~al\mbox{.}(2021)]%
        {DiffusionCLIP}
\bibfield{author}{\bibinfo{person}{Gwanghyun Kim}, \bibinfo{person}{Taesung
  Kwon}, {and} \bibinfo{person}{Jong~Chul Ye}.}
  \bibinfo{year}{2021}\natexlab{}.
\newblock \showarticletitle{DiffusionCLIP: Text-Guided Diffusion Models for
  Robust Image Manipulation}.
\newblock  (\bibinfo{year}{2021}).
\newblock
\urldef\tempurl%
\url{https://doi.org/10.48550/ARXIV.2110.02711}
\showDOI{\tempurl}


\bibitem[Kingma and Dhariwal(2018)]%
        {DiederikPKingma2018GlowGF}
\bibfield{author}{\bibinfo{person}{Diederik~P. Kingma} {and}
  \bibinfo{person}{Prafulla Dhariwal}.} \bibinfo{year}{2018}\natexlab{}.
\newblock \showarticletitle{Glow: Generative Flow with Invertible 1x1
  Convolutions}.
\newblock \bibinfo{journal}{\emph{arXiv: Machine Learning}}
  (\bibinfo{year}{2018}).
\newblock


\bibitem[Kotovenko et~al\mbox{.}(2019)]%
        {kotovenko2019content}
\bibfield{author}{\bibinfo{person}{Dmytro Kotovenko}, \bibinfo{person}{Artsiom
  Sanakoyeu}, \bibinfo{person}{Sabine Lang}, {and} \bibinfo{person}{Bjorn
  Ommer}.} \bibinfo{year}{2019}\natexlab{}.
\newblock \showarticletitle{Content and Style Disentanglement for Artistic
  Style Transfer}. In \bibinfo{booktitle}{\emph{IEEE/CVF International
  Conference on Computer Vision (ICCV)}}. \bibinfo{pages}{4422--4431}.
\newblock


\bibitem[Kwon and Ye(2022)]%
        {CLIPstyler}
\bibfield{author}{\bibinfo{person}{Gihyun Kwon} {and}
  \bibinfo{person}{Jong~Chul Ye}.} \bibinfo{year}{2022}\natexlab{}.
\newblock \showarticletitle{CLIPstyler: Image Style Transfer with a Single Text
  Condition}. In \bibinfo{booktitle}{\emph{IEEE/CVF Conference on Computer
  Vision and Pattern Recognition (CVPR)}}.
\newblock


\bibitem[Li et~al\mbox{.}(2020a)]%
        {diffvg}
\bibfield{author}{\bibinfo{person}{Tzu-Mao Li}, \bibinfo{person}{Michal
  Luk{\'a}{\v{c}}}, \bibinfo{person}{Micha{\"e}l Gharbi}, {and}
  \bibinfo{person}{Jonathan Ragan-Kelley}.} \bibinfo{year}{2020}\natexlab{a}.
\newblock \showarticletitle{Differentiable vector graphics rasterization for
  editing and learning}.
\newblock \bibinfo{journal}{\emph{ACM Transactions on Graphics}}
  \bibinfo{volume}{39}, \bibinfo{number}{6} (\bibinfo{year}{2020}),
  \bibinfo{pages}{1--15}.
\newblock


\bibitem[Li et~al\mbox{.}(2020b)]%
        {li2020oscar}
\bibfield{author}{\bibinfo{person}{Xiujun Li}, \bibinfo{person}{Xi Yin},
  \bibinfo{person}{Chunyuan Li}, \bibinfo{person}{Pengchuan Zhang},
  \bibinfo{person}{Xiaowei Hu}, \bibinfo{person}{Lei Zhang},
  \bibinfo{person}{Lijuan Wang}, \bibinfo{person}{Houdong Hu},
  \bibinfo{person}{Li Dong}, \bibinfo{person}{Furu Wei}, {et~al\mbox{.}}}
  \bibinfo{year}{2020}\natexlab{b}.
\newblock \showarticletitle{Oscar: Object-semantics aligned pre-training for
  vision-language tasks}. In \bibinfo{booktitle}{\emph{European Conference on
  Computer Vision (ECCV)}}. \bibinfo{pages}{121--137}.
\newblock


\bibitem[Lin et~al\mbox{.}(2021)]%
        {Lin:2021:DAM}
\bibfield{author}{\bibinfo{person}{Minxuan Lin}, \bibinfo{person}{Fan Tang},
  \bibinfo{person}{Weiming Dong}, \bibinfo{person}{Xiao Li},
  \bibinfo{person}{Changsheng Xu}, {and} \bibinfo{person}{Chongyang Ma}.}
  \bibinfo{year}{2021}\natexlab{}.
\newblock \showarticletitle{Distribution Aligned Multimodal and Multi-Domain
  Image Stylization}.
\newblock \bibinfo{journal}{\emph{ACM Transactions on Multimedia Computing,
  Communications, and Applications}} \bibinfo{volume}{17}, \bibinfo{number}{3},
  Article \bibinfo{articleno}{96} (\bibinfo{year}{2021}),
  \bibinfo{numpages}{17}~pages.
\newblock
\showISSN{1551-6857}


\bibitem[Liu et~al\mbox{.}(2021b)]%
        {liu2021adaattn}
\bibfield{author}{\bibinfo{person}{Songhua Liu}, \bibinfo{person}{Tianwei Lin},
  \bibinfo{person}{Dongliang He}, \bibinfo{person}{Fu Li},
  \bibinfo{person}{Meiling Wang}, \bibinfo{person}{Xin Li},
  \bibinfo{person}{Zhengxing Sun}, \bibinfo{person}{Qian Li}, {and}
  \bibinfo{person}{Errui Ding}.} \bibinfo{year}{2021}\natexlab{b}.
\newblock \showarticletitle{{AdaAttN}: Revisit attention mechanism in arbitrary
  neural style transfer}. In \bibinfo{booktitle}{\emph{IEEE/CVF International
  Conference on Computer Vision (ICCV)}}. \bibinfo{pages}{6649--6658}.
\newblock


\bibitem[Liu et~al\mbox{.}(2021a)]%
        {fusedream}
\bibfield{author}{\bibinfo{person}{Xingchao Liu}, \bibinfo{person}{Chengyue
  Gong}, \bibinfo{person}{Lemeng Wu}, \bibinfo{person}{Shujian Zhang},
  \bibinfo{person}{Hao Su}, {and} \bibinfo{person}{Qiang Liu}.}
  \bibinfo{year}{2021}\natexlab{a}.
\newblock \bibinfo{title}{FuseDream: Training-Free Text-to-Image Generation
  with Improved CLIP+GAN Space Optimization}.
\newblock
\newblock
\showeprint[arxiv]{2112.01573}~[cs.CV]


\bibitem[Liu et~al\mbox{.}(2021c)]%
        {MoreCF}
\bibfield{author}{\bibinfo{person}{Xihui Liu}, \bibinfo{person}{Dong~Huk Park},
  \bibinfo{person}{Samaneh Azadi}, \bibinfo{person}{Gong Zhang},
  \bibinfo{person}{Arman Chopikyan}, \bibinfo{person}{Yuxiao Hu},
  \bibinfo{person}{Humphrey Shi}, \bibinfo{person}{Anna Rohrbach}, {and}
  \bibinfo{person}{Trevor Darrell}.} \bibinfo{year}{2021}\natexlab{c}.
\newblock \showarticletitle{More Control for Free! Image Synthesis with
  Semantic Diffusion Guidance}.
\newblock
\showeprint[arxiv]{2112.05744}~[cs.CV]


\bibitem[Liu et~al\mbox{.}(2020)]%
        {YahuiLiu2020DescribeWT}
\bibfield{author}{\bibinfo{person}{Yahui Liu}, \bibinfo{person}{Marco
  De~Nadai}, \bibinfo{person}{Deng Cai}, \bibinfo{person}{Huayang Li},
  \bibinfo{person}{Xavier Alameda-Pineda}, \bibinfo{person}{Nicu Sebe}, {and}
  \bibinfo{person}{Bruno Lepri}.} \bibinfo{year}{2020}\natexlab{}.
\newblock \showarticletitle{Describe What to Change: A Text-Guided Unsupervised
  Image-to-Image Translation Approach}. In
  \bibinfo{booktitle}{\emph{Proceedings of the 28th ACM International
  Conference on Multimedia}}. \bibinfo{publisher}{Association for Computing
  Machinery}, \bibinfo{pages}{1357–1365}.
\newblock


\bibitem[Menick and Kalchbrenner(2018)]%
        {JacobMenick2018GeneratingHF}
\bibfield{author}{\bibinfo{person}{Jacob Menick} {and} \bibinfo{person}{Nal
  Kalchbrenner}.} \bibinfo{year}{2018}\natexlab{}.
\newblock \showarticletitle{Generating High Fidelity Images with Subscale Pixel
  Networks and Multidimensional Upscaling}.
\newblock \bibinfo{journal}{\emph{international conference on learning
  representations (ICLR)}} (\bibinfo{year}{2018}).
\newblock


\bibitem[Nichol et~al\mbox{.}(2021)]%
        {glide}
\bibfield{author}{\bibinfo{person}{Alex Nichol}, \bibinfo{person}{Prafulla
  Dhariwal}, \bibinfo{person}{Aditya Ramesh}, \bibinfo{person}{Pranav Shyam},
  \bibinfo{person}{Pamela Mishkin}, \bibinfo{person}{Bob McGrew},
  \bibinfo{person}{Ilya Sutskever}, {and} \bibinfo{person}{Mark Chen}.}
  \bibinfo{year}{2021}\natexlab{}.
\newblock \showarticletitle{Glide: Towards photorealistic image generation and
  editing with text-guided diffusion models}.
\newblock \bibinfo{journal}{\emph{arXiv preprint arXiv:2112.10741}}
  (\bibinfo{year}{2021}).
\newblock


\bibitem[Patashnik et~al\mbox{.}(2021)]%
        {StyleCLIP}
\bibfield{author}{\bibinfo{person}{Or Patashnik}, \bibinfo{person}{Zongze Wu},
  \bibinfo{person}{Eli Shechtman}, \bibinfo{person}{Daniel Cohen-Or}, {and}
  \bibinfo{person}{Dani Lischinski}.} \bibinfo{year}{2021}\natexlab{}.
\newblock \showarticletitle{StyleCLIP: Text-Driven Manipulation of StyleGAN
  Imagery}. In \bibinfo{booktitle}{\emph{Proceedings of the IEEE/CVF
  International Conference on Computer Vision (ICCV)}}.
  \bibinfo{pages}{2085--2094}.
\newblock


\bibitem[Radford et~al\mbox{.}(2021)]%
        {openai}
\bibfield{author}{\bibinfo{person}{Alec Radford}, \bibinfo{person}{Jong~Wook
  Kim}, \bibinfo{person}{Chris Hallacy}, \bibinfo{person}{Aditya Ramesh},
  \bibinfo{person}{Gabriel Goh}, \bibinfo{person}{Sandhini Agarwal},
  \bibinfo{person}{Girish Sastry}, \bibinfo{person}{Amanda Askell},
  \bibinfo{person}{Pamela Mishkin}, \bibinfo{person}{Jack Clark},
  {et~al\mbox{.}}} \bibinfo{year}{2021}\natexlab{}.
\newblock \showarticletitle{Learning transferable visual models from natural
  language supervision}. In \bibinfo{booktitle}{\emph{International Conference
  on Machine Learning}}. PMLR, \bibinfo{pages}{8748--8763}.
\newblock


\bibitem[Ramesh et~al\mbox{.}(2022)]%
        {dalle2}
\bibfield{author}{\bibinfo{person}{Aditya Ramesh}, \bibinfo{person}{Prafulla
  Dhariwal}, \bibinfo{person}{Alex Nichol}, \bibinfo{person}{Casey Chu}, {and}
  \bibinfo{person}{Mark Chen}.} \bibinfo{year}{2022}\natexlab{}.
\newblock \showarticletitle{Hierarchical text-conditional image generation with
  clip latents}.
\newblock \bibinfo{journal}{\emph{arXiv preprint arXiv:2204.06125}}
  (\bibinfo{year}{2022}).
\newblock


\bibitem[Ramesh et~al\mbox{.}(2021)]%
        {AdityaRamesh2021ZeroShotTG}
\bibfield{author}{\bibinfo{person}{Aditya Ramesh}, \bibinfo{person}{Mikhail
  Pavlov}, \bibinfo{person}{Gabriel Goh}, \bibinfo{person}{Scott Gray},
  \bibinfo{person}{Chelsea Voss}, \bibinfo{person}{Alec Radford},
  \bibinfo{person}{Mark Chen}, {and} \bibinfo{person}{Ilya Sutskever}.}
  \bibinfo{year}{2021}\natexlab{}.
\newblock \showarticletitle{Zero-Shot Text-to-Image Generation}. In
  \bibinfo{booktitle}{\emph{International Conference on Machine Learning
  (ICML)}}. \bibinfo{pages}{8821--8831}.
\newblock


\bibitem[Razavi et~al\mbox{.}(2019)]%
        {razavi2019generating}
\bibfield{author}{\bibinfo{person}{Ali Razavi}, \bibinfo{person}{A\"aron
  van~den Oord}, {and} \bibinfo{person}{Oriol Vinyals}.}
  \bibinfo{year}{2019}\natexlab{}.
\newblock \showarticletitle{Generating Diverse High-Fidelity Images with
  VQ-VAE-2}. In \bibinfo{booktitle}{\emph{Advances in Neural Information
  Processing Systems}}.
\newblock


\bibitem[Rodent(2022)]%
        {vqgan-clip}
\bibfield{author}{\bibinfo{person}{Nerdy Rodent}.}
  \bibinfo{year}{2022}\natexlab{}.
\newblock \bibinfo{booktitle}{\emph{Source Code of VQGAN-CLIP}}.
\newblock
\urldef\tempurl%
\url{https://github.com/nerdyrodent/VQGAN-CLIP}
\showURL{%
\tempurl}


\bibitem[Rombach et~al\mbox{.}(2021)]%
        {latentDiffusion}
\bibfield{author}{\bibinfo{person}{Robin Rombach}, \bibinfo{person}{Andreas
  Blattmann}, \bibinfo{person}{Dominik Lorenz}, \bibinfo{person}{Patrick
  Esser}, {and} \bibinfo{person}{Bj\"orn Ommer}.}
  \bibinfo{year}{2021}\natexlab{}.
\newblock \showarticletitle{High-Resolution Image Synthesis with Latent
  Diffusion Models}.
\newblock


\bibitem[Ronneberger et~al\mbox{.}(2015)]%
        {UNet}
\bibfield{author}{\bibinfo{person}{Olaf Ronneberger}, \bibinfo{person}{Philipp
  Fischer}, {and} \bibinfo{person}{Thomas Brox}.}
  \bibinfo{year}{2015}\natexlab{}.
\newblock \showarticletitle{U-Net: Convolutional Networks for Biomedical Image
  Segmentation}. In \bibinfo{booktitle}{\emph{Medical Image Computing and
  Computer-Assisted Intervention}}.
\newblock


\bibitem[Ruan et~al\mbox{.}(2021)]%
        {ruan2021dae}
\bibfield{author}{\bibinfo{person}{Shulan Ruan}, \bibinfo{person}{Yong Zhang},
  \bibinfo{person}{Kun Zhang}, \bibinfo{person}{Yanbo Fan},
  \bibinfo{person}{Fan Tang}, \bibinfo{person}{Qi Liu}, {and}
  \bibinfo{person}{Enhong Chen}.} \bibinfo{year}{2021}\natexlab{}.
\newblock \showarticletitle{{DAE-GAN}: Dynamic Aspect-aware {GAN} for
  Text-to-Image Synthesis}. In \bibinfo{booktitle}{\emph{IEEE/CVF International
  Conference on Computer Vision (ICCV)}}. \bibinfo{pages}{13960--13969}.
\newblock


\bibitem[Schaldenbrand et~al\mbox{.}(2022)]%
        {styleclipdraw}
\bibfield{author}{\bibinfo{person}{Peter Schaldenbrand},
  \bibinfo{person}{Zhixuan Liu}, {and} \bibinfo{person}{Jean Oh}.}
  \bibinfo{year}{2022}\natexlab{}.
\newblock \showarticletitle{StyleCLIPDraw: Coupling Content and Style in
  Text-to-Drawing Translation}.
\newblock \bibinfo{journal}{\emph{arXiv preprint arXiv:2202.12362}}
  (\bibinfo{year}{2022}).
\newblock


\bibitem[Sohl-Dickstein et~al\mbox{.}(2015a)]%
        {2015Diffusion}
\bibfield{author}{\bibinfo{person}{Jascha Sohl-Dickstein},
  \bibinfo{person}{Eric~L. Weiss}, \bibinfo{person}{Niru Maheswaranathan},
  {and} \bibinfo{person}{Surya Ganguli}.} \bibinfo{year}{2015}\natexlab{a}.
\newblock \showarticletitle{Deep Unsupervised Learning using Nonequilibrium
  Thermodynamics}.
\newblock \bibinfo{journal}{\emph{arXiv: Learning}} (\bibinfo{year}{2015}).
\newblock


\bibitem[Sohl-Dickstein et~al\mbox{.}(2015b)]%
        {JaschaSohlDickstein2015DeepUL}
\bibfield{author}{\bibinfo{person}{Jascha Sohl-Dickstein},
  \bibinfo{person}{Eric~L. Weiss}, \bibinfo{person}{Niru Maheswaranathan},
  {and} \bibinfo{person}{Surya Ganguli}.} \bibinfo{year}{2015}\natexlab{b}.
\newblock \showarticletitle{Deep Unsupervised Learning using Nonequilibrium
  Thermodynamics}.
\newblock \bibinfo{journal}{\emph{arXiv: Learning}} (\bibinfo{year}{2015}).
\newblock


\bibitem[Song et~al\mbox{.}(2021a)]%
        {DDIM}
\bibfield{author}{\bibinfo{person}{Jiaming Song}, \bibinfo{person}{Chenlin
  Meng}, {and} \bibinfo{person}{Stefano Ermon}.}
  \bibinfo{year}{2021}\natexlab{a}.
\newblock \showarticletitle{Denoising Diffusion Implicit Models}. In
  \bibinfo{booktitle}{\emph{International Conference on Learning
  Representations (ICLR)}}.
\newblock


\bibitem[Song and Ermon(2019)]%
        {song2019generative}
\bibfield{author}{\bibinfo{person}{Yang Song} {and} \bibinfo{person}{Stefano
  Ermon}.} \bibinfo{year}{2019}\natexlab{}.
\newblock \showarticletitle{Generative modeling by estimating gradients of the
  data distribution}.
\newblock \bibinfo{journal}{\emph{Advances in Neural Information Processing
  Systems}}  \bibinfo{volume}{32} (\bibinfo{year}{2019}).
\newblock


\bibitem[Song et~al\mbox{.}(2021b)]%
        {YangSong2021ScoreBasedGM}
\bibfield{author}{\bibinfo{person}{Yang Song}, \bibinfo{person}{Jascha
  Sohl-Dickstein}, \bibinfo{person}{Diederik~P. Kingma},
  \bibinfo{person}{Abhishek Kumar}, \bibinfo{person}{Stefano Ermon}, {and}
  \bibinfo{person}{Ben Poole}.} \bibinfo{year}{2021}\natexlab{b}.
\newblock \showarticletitle{Score-Based Generative Modeling through Stochastic
  Differential Equations}. In \bibinfo{booktitle}{\emph{International
  Conference on Learning Representations (ICLR)}}.
\newblock


\bibitem[Tan et~al\mbox{.}(2017)]%
        {ArtGAN}
\bibfield{author}{\bibinfo{person}{Wei~Ren Tan}, \bibinfo{person}{Chee~Seng
  Chan}, \bibinfo{person}{Hern{\'a}n Aguirre}, {and} \bibinfo{person}{Kiyoshi
  Tanaka}.} \bibinfo{year}{2017}\natexlab{}.
\newblock \showarticletitle{ArtGAN: Artwork Synthesis with Conditional
  Categorial GANs.}
\newblock


\bibitem[Tan et~al\mbox{.}(2019)]%
        {ImprovedArtGAN}
\bibfield{author}{\bibinfo{person}{Wei~Ren Tan}, \bibinfo{person}{Chee~Seng
  Chan}, \bibinfo{person}{Hern{\'a}n Aguirre}, {and} \bibinfo{person}{Kiyoshi
  Tanaka}.} \bibinfo{year}{2019}\natexlab{}.
\newblock \showarticletitle{Improved ArtGAN for Conditional Synthesis of
  Natural Image and Artwork}.
\newblock \bibinfo{journal}{\emph{IEEE Transactions on Image Processing}}
  \bibinfo{volume}{28} (\bibinfo{year}{2019}), \bibinfo{pages}{394--409}.
\newblock


\bibitem[Tang et~al\mbox{.}(2018)]%
        {tang2017animated}
\bibfield{author}{\bibinfo{person}{Fan Tang}, \bibinfo{person}{Weiming Dong},
  \bibinfo{person}{Yiping Meng}, \bibinfo{person}{Xing Mei},
  \bibinfo{person}{Feiyue Huang}, \bibinfo{person}{Xiaopeng Zhang}, {and}
  \bibinfo{person}{Oliver Deussen}.} \bibinfo{year}{2018}\natexlab{}.
\newblock \showarticletitle{Animated Construction of Chinese Brush Paintings}.
\newblock \bibinfo{journal}{\emph{IEEE Transactions on Visualization and
  Computer Graphics}} \bibinfo{volume}{24}, \bibinfo{number}{12}
  (\bibinfo{year}{2018}), \bibinfo{pages}{3019--3031}.
\newblock


\bibitem[Wang et~al\mbox{.}(2021)]%
        {HaoWang2021CycleConsistentIG}
\bibfield{author}{\bibinfo{person}{Hao Wang}, \bibinfo{person}{Guosheng Lin},
  \bibinfo{person}{Steven C.~H. Hoi}, {and} \bibinfo{person}{Chunyan Miao}.}
  \bibinfo{year}{2021}\natexlab{}.
\newblock \showarticletitle{Cycle-Consistent Inverse GAN for Text-to-Image
  Synthesis}. In \bibinfo{booktitle}{\emph{Proceedings of the 29th ACM
  International Conference on Multimedia}}. \bibinfo{publisher}{Association for
  Computing Machinery}, \bibinfo{address}{New York, NY, USA},
  \bibinfo{pages}{630–638}.
\newblock


\bibitem[Wang et~al\mbox{.}(2014)]%
        {Wang:2014:TPW}
\bibfield{author}{\bibinfo{person}{Miaoyi Wang}, \bibinfo{person}{Bin Wang},
  \bibinfo{person}{Yun Fei}, \bibinfo{person}{Kanglai Qian},
  \bibinfo{person}{Wenping Wang}, \bibinfo{person}{Jiating Chen}, {and}
  \bibinfo{person}{Jun-Hai Yong}.} \bibinfo{year}{2014}\natexlab{}.
\newblock \showarticletitle{Towards Photo Watercolorization with Artistic
  Verisimilitude}.
\newblock \bibinfo{journal}{\emph{IEEE Transactions on Visualization and
  Computer Graphics}} \bibinfo{volume}{20}, \bibinfo{number}{10}
  (\bibinfo{year}{2014}), \bibinfo{pages}{1451--1460}.
\newblock


\bibitem[Wang et~al\mbox{.}(2020)]%
        {Wang:2020:DAS}
\bibfield{author}{\bibinfo{person}{Zhizhong Wang}, \bibinfo{person}{Lei Zhao},
  \bibinfo{person}{Haibo Chen}, \bibinfo{person}{Lihong Qiu},
  \bibinfo{person}{Qihang Mo}, \bibinfo{person}{Sihuan Lin},
  \bibinfo{person}{Wei Xing}, {and} \bibinfo{person}{Dongming Lu}.}
  \bibinfo{year}{2020}\natexlab{}.
\newblock \showarticletitle{Diversified Arbitrary Style Transfer via Deep
  Feature Perturbation}. In \bibinfo{booktitle}{\emph{IEEE/CVF Conference on
  Computer Vision and Pattern Recognition (CVPR)}}.
  \bibinfo{pages}{7786--7795}.
\newblock


\bibitem[Wei et~al\mbox{.}(2022)]%
        {Wei:2022:ACS}
\bibfield{author}{\bibinfo{person}{Hua-Peng Wei}, \bibinfo{person}{Ying-Ying
  Deng}, \bibinfo{person}{Fan Tang}, \bibinfo{person}{Xing-Jia Pan}, {and}
  \bibinfo{person}{Wei-Ming Dong}.} \bibinfo{year}{2022}\natexlab{}.
\newblock \showarticletitle{A Comparative Study of CNN- and Transformer-Based
  Visual Style Transfer}.
\newblock \bibinfo{journal}{\emph{Journal of Computer Science and Technology}}
  \bibinfo{volume}{37}, \bibinfo{number}{3} (\bibinfo{year}{2022}),
  \bibinfo{pages}{601–614}.
\newblock


\bibitem[Xu et~al\mbox{.}(2018)]%
        {TaoXu2018AttnGANFT}
\bibfield{author}{\bibinfo{person}{Tao Xu}, \bibinfo{person}{Pengchuan Zhang},
  \bibinfo{person}{Qiuyuan Huang}, \bibinfo{person}{Han Zhang},
  \bibinfo{person}{Zhe Gan}, \bibinfo{person}{Xiaolei Huang}, {and}
  \bibinfo{person}{Xiaodong He}.} \bibinfo{year}{2018}\natexlab{}.
\newblock \showarticletitle{AttnGAN: Fine-Grained Text to Image Generation with
  Attentional Generative Adversarial Networks}. In
  \bibinfo{booktitle}{\emph{IEEE/CVF Conference on Computer Vision and Pattern
  Recognition (CVPR)}}. \bibinfo{pages}{1316--1324}.
\newblock


\bibitem[Xue et~al\mbox{.}(2022)]%
        {xue2022deep}
\bibfield{author}{\bibinfo{person}{Yuan Xue}, \bibinfo{person}{Yuan-Chen Guo},
  \bibinfo{person}{Han Zhang}, \bibinfo{person}{Tao Xu},
  \bibinfo{person}{Song-Hai Zhang}, {and} \bibinfo{person}{Xiaolei Huang}.}
  \bibinfo{year}{2022}\natexlab{}.
\newblock \showarticletitle{Deep image synthesis from intuitive user input: A
  review and perspectives}.
\newblock \bibinfo{journal}{\emph{Computational Visual Media}}
  \bibinfo{volume}{8}, \bibinfo{number}{1} (\bibinfo{year}{2022}),
  \bibinfo{pages}{3--31}.
\newblock


\bibitem[Yi et~al\mbox{.}(2021)]%
        {PaintingDiffusion}
\bibfield{author}{\bibinfo{person}{Da Yi}, \bibinfo{person}{Chao Guo}, {and}
  \bibinfo{person}{Tianxiang Bai}.} \bibinfo{year}{2021}\natexlab{}.
\newblock \showarticletitle{Exploring Painting Synthesis with Diffusion
  Models}. In \bibinfo{booktitle}{\emph{IEEE 1st International Conference on
  Digital Twins and Parallel Intelligence (DTPI)}}. \bibinfo{pages}{332--335}.
\newblock


\bibitem[Zagoruyko and Komodakis(2016)]%
        {wideres}
\bibfield{author}{\bibinfo{person}{Sergey Zagoruyko} {and}
  \bibinfo{person}{Nikos Komodakis}.} \bibinfo{year}{2016}\natexlab{}.
\newblock \showarticletitle{Wide residual networks}.
\newblock \bibinfo{journal}{\emph{arXiv preprint arXiv:1605.07146}}
  (\bibinfo{year}{2016}).
\newblock


\bibitem[Zhang and Yu(2016)]%
        {Zhang:2016:GKA}
\bibfield{author}{\bibinfo{person}{Kang Zhang} {and} \bibinfo{person}{Jinhui
  Yu}.} \bibinfo{year}{2016}\natexlab{}.
\newblock \showarticletitle{Generation of Kandinsky Art}.
\newblock \bibinfo{journal}{\emph{Leonardo}} \bibinfo{volume}{49},
  \bibinfo{number}{1} (\bibinfo{year}{2016}), \bibinfo{pages}{48--54}.
\newblock


\bibitem[Zhang et~al\mbox{.}(2018)]%
        {LPIPS}
\bibfield{author}{\bibinfo{person}{Richard Zhang}, \bibinfo{person}{Phillip
  Isola}, \bibinfo{person}{Alexei~A Efros}, \bibinfo{person}{Eli Shechtman},
  {and} \bibinfo{person}{Oliver Wang}.} \bibinfo{year}{2018}\natexlab{}.
\newblock \showarticletitle{The unreasonable effectiveness of deep features as
  a perceptual metric}. In \bibinfo{booktitle}{\emph{Proceedings of the IEEE
  Conference on Computer Vision and Pattern Recognition (CVPR)}}.
  \bibinfo{pages}{586--595}.
\newblock


\bibitem[Zhang et~al\mbox{.}(2022)]%
        {Zhang:2022:CAST}
\bibfield{author}{\bibinfo{person}{Yuxin Zhang}, \bibinfo{person}{Fan Tang},
  \bibinfo{person}{Weiming Dong}, \bibinfo{person}{Haibin Huang},
  \bibinfo{person}{Chongyang Ma}, \bibinfo{person}{Tong-Yee Lee}, {and}
  \bibinfo{person}{Changsheng Xu}.} \bibinfo{year}{2022}\natexlab{}.
\newblock \showarticletitle{Domain Enhanced Arbitrary Image Style Transfer via
  Contrastive Learning}. In \bibinfo{booktitle}{\emph{Special Interest Group on
  Computer Graphics and Interactive Techniques Conference Proceedings (SIGGRAPH
  ’22 Conference Proceedings}}.
\newblock


\bibitem[Zhao et~al\mbox{.}(2018)]%
        {BoZhao2018MultiViewIG}
\bibfield{author}{\bibinfo{person}{Bo Zhao}, \bibinfo{person}{Xiao Wu},
  \bibinfo{person}{Zhi-Qi Cheng}, \bibinfo{person}{Hao Liu},
  \bibinfo{person}{Zequn Jie}, {and} \bibinfo{person}{Jiashi Feng}.}
  \bibinfo{year}{2018}\natexlab{}.
\newblock \showarticletitle{Multi-View Image Generation from a Single-View}. In
  \bibinfo{booktitle}{\emph{Proceedings of the 26th ACM International
  Conference on Multimedia}} (Seoul, Republic of Korea).
  \bibinfo{publisher}{Association for Computing Machinery},
  \bibinfo{address}{New York, NY, USA}, \bibinfo{pages}{383–391}.
\newblock
\showISBNx{9781450356657}


\bibitem[Zhu et~al\mbox{.}(2017)]%
        {Zhu:2017:CycleGAN}
\bibfield{author}{\bibinfo{person}{Jun-Yan Zhu}, \bibinfo{person}{Taesung
  Park}, \bibinfo{person}{Phillip Isola}, {and} \bibinfo{person}{Alexei~A
  Efros}.} \bibinfo{year}{2017}\natexlab{}.
\newblock \showarticletitle{Unpaired Image-to-Image Translation Using
  Cycle-Consistent Adversarial Networks}. In
  \bibinfo{booktitle}{\emph{Proceedings of the IEEE International Conference on
  Computer Vision (ICCV)}}. \bibinfo{pages}{2223--2232}.
\newblock


\end{thebibliography}

\end{document}